\documentclass[journal]{IEEEtran}
\usepackage{graphics} % for pdf, bitmapped graphics files
\usepackage{epsfig} % for postscript graphics files
\usepackage{mathptmx} % assumes new font selection scheme installed
\usepackage{amsmath} % assumes amsmath package installed
\usepackage[utf8]{inputenc}
\usepackage[T1]{fontenc}
\usepackage{mathtools}
\usepackage{enumitem}
\usepackage{xcolor}
\usepackage{float}
\usepackage{gensymb}
\usepackage{bm}
\usepackage{autobreak}
\usepackage{svg}
\usepackage{mathrsfs}
\usepackage{tabularx}
\usepackage{array, makecell}
\usepackage{graphicx}
\usepackage{cellspace}
\usepackage{textcomp}
\usepackage{hyperref}
\usepackage{cleveref}
\usepackage{booktabs}
\usepackage{algorithm}
\usepackage{algpseudocode}
\usepackage{blkarray}
\usepackage{xcolor}
\usepackage{comment}

\newcommand{\cusst}[1]{{}}

\DeclareMathAlphabet{\mathcal}{OMS}{cmsy}{m}{n}  % thus can use mathcal{}
\algnewcommand{\LineComment}[1]{\State \(\triangleright\) #1}

\begin{document}

\title{Toward Extending Concentric Tube Robot Kinematics for Large Clearance and Impulse Curvature
} 

 \author{Zhouyu~Zhang*, Jia~Shen*, Junhyoung~Ha, and Yue~Chen% <-this % stops a space
 \thanks{Research reported in this publication was supported the Georgia Tech Faculty Startup Grant and  McCamish Blue Sky Research Program. It was also partially supported  by the National Institute of Biomedical Imaging And Bioengineering of the National Institutes of Health under Award Number R01EB034359. The content is solely the responsibility of the authors and does not necessarily represent the official views of the National Institutes of Health.  Corresponding author: Yue Chen. }
 \thanks{J. Shen and Z. Zhang are with the Department of Mechanical Engineering, Georgia Institute of Technology, Atlanta 30332 USA (e-mail: \{jshen359, zzhang3097\}@gatech.edu.)}% <-this % stops a space

\thanks{Junhyoung Ha is with the Center for Healthcare Robotics, Artificial Intelligence
and Robotics Institute, Korea Institute of Science and Technology, Seoul 02792,
South Korea (e-mail: jhha@kist.re.kr)}

 \thanks{Y. Chen is with the Wallace H. Coulter  Department of Biomedical Engineering, Georgia Institute of Technology/Emory University, Atlanta 30332 USA (e-mail: yue.chen@bme.gatech.edu)}

 \thanks{* Zhouyu Zhang and Jia Shen contributed equally to this paper. }
 \thanks{This work has been submitted to IEEE for possible publication. Copyright may be transferred without notice, after which the version may no longer be accessible.}}

% % <-this % stops a space
% % \thanks{Manuscript received April 19, 2005; revised August 26, 2015.}
% }


\markboth{}%IEEE Robotics and Automation Letters
{Shell \MakeLowercase{\textit{et al.}}: Bare Demo of IEEEtran.cls for IEEE Journals}



\maketitle

\begin{abstract}
Concentric Tube Robots (CTRs) have been proposed to operate within the unstructured environment for minimally invasive surgeries.  In this letter, we consider the operation scenario where the tubes travel inside the channels with a large clearance or large curvature, such as aortas  or industrial pipes. Accurate kinematic modeling of CTRs is required for the development of advanced control and sensing algorithms.  To this end, we extended the conventional CTR kinematics model to a more general case with large tube-to-tube clearance and large centerline curvature. Numerical simulations and experimental validations are conducted to compare our model with respect to the conventional CTR kinematic model. In the physical experiments, our proposed model achieved a tip position error of 1.53 mm in the 2D planer case and 4.36 mm in 3D case, outperforming the state-of-the-art model by 71\% and 66\%, respectively. 
\end{abstract}

% Note that keywords are not normally used for peer review papers.

% \yc{let's try to get a system application figure and put in page 1. for example, get blood vessel model from here: \url{https://www.youtube.com/watch?v=yi07mjr3JeU}, and then perform simulation studies to show the difference between the proposed model and current model}
% \fi
\IEEEpeerreviewmaketitle

\section{Introduction}

Concentric tube robots (CTRs) are a class of continuum robots that have gained significant attention and progress in various fields, particularly in surgery and interventional medicine \cite{dupont2022continuum}\cite{kim2015cardioscopic}\cite{butler2012robotic}. These robots are composed of multiple concentrically arranged super-elastic tubes, with each tube capable of independent actuation. By selectively controlling the linear translation and axial rotation of the tubes, CTRs can reconfigure themselves and generate tentacle-like motions to navigate through complex anatomical structures. % with enhanced maneuverability and dexterity.

Accurate modeling   of CTRs is of paramount importance for reliable manipulation. % since it serves as premise for the optimization of control algorithms and insights into the robot's workspace, limitations, and capabilities. 
 Cosserat rod theory has been extensively used to describe the kinematics of CTRs \cite{dupont2009design}\cite{dupont2009torsional}\cite{geoexact}. Prior work indicated that the kinematics  of CTRs involves formulating a set of differential equations with mixed boundary conditions \cite{Dupont_2010_CTR_model}, which are derived based on the idea of energy minimization or moment equilibrium. The boundary conditions at the base of the robot are the axial rotations and translations between the tubes, and the boundary conditions at the tip are vanishing internal moments. Numerical methods, such as the shooting method, have been employed to solve these differential equations. Additionally, the kinematic models have been extended to provide differential kinematics, such as inverse kinematics and Jacobian matrices, enabling real-time control of CTRs \cite{dupont2010real}\cite{rucker2011computing}\cite{task_space_position_control}. Recent efforts have also been put into the investigation of CTRs with external loading \cite{geoexact}. 
 
The presence of tube clearance can significantly affect the accuracy of CTRs as the assumption of concentricity may no longer hold true \cite{geoexact}. 
A recent numerical study indicated that the clearence could contribute up to 38.11\% tip error when a miniature continuum robot deployed within a trocar that has a clearence of 0.25mm \cite{ding2022incorporating}. 
%{The model has seen application in the kinematic analysis of CTRs deployed with narrow trocars as guiding cannula \cite{ding2022incorporating}, reducing the tip error by 38.11\%.}
In 2017, co-author Dr. Ha took a pioneering step towards enhancing CTR modeling accuracy by incorporating the  intertube clearance  effect \cite{ha_2017, ha2018modeling}. 
Recently,  \cite{liu2023modeling} proposed the database method to compensate the error caused the intertube clearance. 
However, both approaches were evaluated with CTRs that have a relatively small clearance and smooth curvature. % we developed and validated the kinematic modeling of CTR that has a relatively small clearance and smooth curvature.   % \yc{this sentence seems good, but i quickly scanned the paper and didn't find their use the same method like Dr. ha's work. }

% \zhouyu{I searched over later works that cited Dr Ha's 2017 and 2019 papers, most of the citation were seen in the introduction part, as an example of advanced modeling of CTRs, while the main contents focused on something else. Some were using this to explain their experimental error (without actual calculation). A few used the Cosserat model elaborated in these two papers. Only 2 in around 50 of the citing papers involved the math of clearance modeling. One paper used small clearance model to enhance the modeling of tip accuracy for CTRs placed in a narrow 0.25mm wide trocar \cite{ding2022incorporating} (by the way, they are from the lab Yifan did his master). Another MDPI paper \cite{liu2023modeling} considered intra tube small clearance, friction, and torsion, using database to compensate the error, but I am not sure if it's necessary to cite it, since they actually used a completely different, data driven method to depict the clearance effects of their specific CTR. Both paper did not push forward the theories, but I decided to cite the first one as their application seems successful.}

In this letter, we present the latest results that extend upon the existing small clearance CTR kinematics model proposed in \cite{ha_2017, ha2018modeling} towards more intricate yet practical scenarios, which involve both large clearance and impulse curvature (see definition in Sec. \ref{preliminaries} ). This type of configuration  is commonly observed in medical and engineering applications. For example, in medicine, a $\sim$2mm  continuum catheter is maneuvered within the  aorta  that has a diameter of about 3cm \cite{hager2002diameters}. In engineering applications, the continuum borescope is able to travel within the pipes that have a significantly larger clearance and impulse curvature on given locations (due to the pipe-pipe intersections with 90-degree elbows).  
%for non-destructive inspection. In addition to the large diameter, the intersected pipes are typically connected with  90-degree elbows (a.k.a., large curvature). 
The models in both operation scenarios can be approximated as the CTRs that have large clearance impulse curvature, where the aorta and pipe can be considered as the outer tube, and the continuum robot operate within can be considered as inner tube. Thus, the development of the CTR modeling 
with these considerations can facilitate the ubiquitous use {of} continuum robot in the more practical scenarios where the conventional CTR  model fails, which forms the basic motivation of the proposed work. 
%This will allow us to model the continuum robot manipulation within a large clearance and/or large curvature using the CTR theory by extending the contr The robot modeling in both scenarios can be considered as the  , we will  consider the aorta or pipe as the outer tube, and can leverage the CTR theory for 

%intersecting pipes, typically assembled using angled elbows or fittings bending at angles like 90 degree. When conducting non-destructive pipe inspection using a CTR mounted camera, it could become crucial to develop an accurate method for calculating the shape of the tube and the position of its tip. This, in turn, enhances the control of the CTR device and improves the accuracy of image reconstruction.

The rest of this letter is formulated in the following structure. Part \ref{preliminaries} presents the mathematical modeling of conventional zero clearance CTR,  large clearance tube, and impulse curvature tube, respectively. In part \ref{numercial_experiments} and \ref{hardware_experiments} the models are validated both in simulations and experimental validations, followed by the conclusion in section \ref{conclusion}. %, a conclusion is presented.

\section{CTR Kinematic Modeling} \label{preliminaries}
In this section, the conventional modeling of CTR is briefly discussed in Sec. \ref{sec_SCM}, and then extended by relaxing the requirement on tube-to-tube clearance and center line curvature. The modeling of the latter two cases can be combined, resulting in a large-clearance-impulse-curvature model (LCIC) to be discussed in  Sec. \ref{sec_Large_clearance} and Sec. \ref{sec_Large_curvature}.

%The term ``adaptive'' emphasizes that the constraints will be updated adaptively during the energy minimization process when solving for CTR kinematics, while the conventional zero or small clearance models have a fixed constraint, as detailed below.
%\zhouyu{In the context of this letter, we consider a frictionless environment to underscore our focus on modeling contact and clearance.}
% \yc{the adaptive term here might be better replaced by iterative, since you iteratively update the constraints/clearance} 

% when solving CTR kinematics through energy minimization, adaptive constraints are adopted in contrast with the zero or small clearance model, which will be discussed in detail in later sections. 

\subsection{Standard Kinematics Model of CTRs} \label{sec_SCM}
The kinematics of a  single tube within the CTR labeled with index $i$ can be obtained from the widely studied Cosserat rod model \cite{geoexact}, which describes the evolution of the position and orientation by a system of ODEs. When concentricity constraint is ensured, the following ODEs  (\ref{p_u_relation}) % of each tube present in CTR 
could be solved %for energy minimization or moment equilibrium
with boundary constraints imposed upon tip and base:
\begin{equation} \label{p_u_relation}
\begin{aligned}
    \mathbf{p}^{\prime}_i(s) & = \mathbf{R}_i(s) \mathbf{e}_3 \\
    \mathbf{R}^{\prime}_i(s) & = \mathbf{R}_i(s) \left[ \mathbf{u}_i(s) \right]_{\times}, \\
\end{aligned}
\end{equation}
where $\mathbf{e}_3 = [0~ 0~ 1]^T$, the prime operator $(\cdot)^{\prime}$ denotes the derivative w.r.t. arc-length $s$, $\mathbf{p}_i(s)$, $\mathbf{R}_i(s)$, and $\mathbf{u}_i(s)$ are the centerline position, material coordinate frame, and 3D curvature vector of tube $i$ at $s$, respectively, and $[\cdot]_{\times}$ represents the mapping from a 3D vector to the corresponding skew-symmetric matrix, i.e., the Lie algebra associated with the special orthogonal group $\text{SO}(3)$. The curvature vector $\mathbf{u}_i(s)$ is determined by the elastic interaction between tubes. In the conventional kinematic models with strict concentricity assumption, the centerlines are identical between all tubes, i.e., $\mathbf{p}(s) = \mathbf{p}_1(s) = \mathbf{p}_2(s) = \cdots$. We refer the readers to \cite{geoexact} for more details.

\subsection{Modeling of CTRs with Large Clearance }
\label{sec_Large_clearance}
\begin{figure}[t!]
    \centering
    \includegraphics[width = \linewidth]{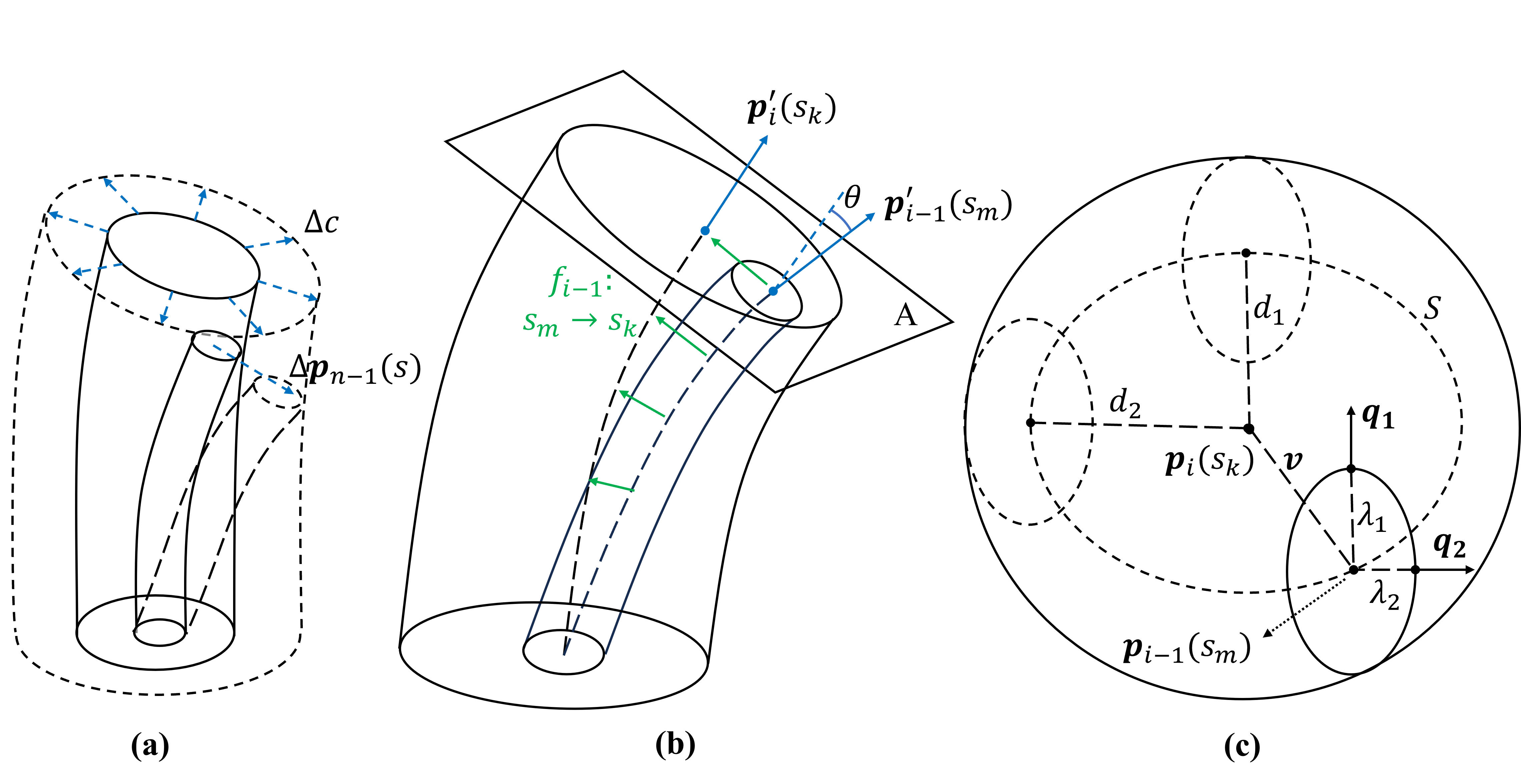}
    \caption{Geometric relations between two adjacent tubes. (a) The change of the inner tube position as the clearance increases even the actuator-space translation and rotation remain constant. (b) The tangent vectors of the tube centerlines at a common cross-section. The mapping $f$ is also displayed (in green) that finds the corresponding points on the outer tube closest to the inner tube. (c) A detailed illustration of the cross-section plane, labeled as plane $A$, that is in perpendicular to the outer tube, }
    \label{fig:cross-section}
    \vspace{- 4mm}
\end{figure}

In the practical implementations of CTRs, the clearances between tubes are required to enable the relative rotation and translation with respect to each other. A more accurate model of CTRs can be obtained by relaxing the concentricity constraint (see Fig. \ref{fig:cross-section}-(a)). In this case, the centerlines of the tubes are no longer identical and will be obtained below. %because the inner tube can shape within the interior space of the outer tube.

To formulate the kinematics as a finite-dimensional problem, the centerline position and curvature are discretized along the arc length {into virtual segments} for each tube. The state variables of tube $i$ are the concatenated vectors of discrete variables, given as
\begin{equation}
    \mathbf{u}^i = [\mathbf{u}_i(s_1)^T, \ldots, \mathbf{u}_i(s_{N_i})^T]^T ,~~~ \mathbf{p}^i = [\mathbf{p}_i(s_1)^T, ..., \mathbf{p}_i(s_{N_i})^T]^T,
\end{equation}
where $s_j$ is the arc length of the $j$-th discretization point along the tube, and $N_i$ is the total number of discretization. 
The state variables of the whole system is defined as concatenation of each tube variables
\begin{equation}
    \mathbf{u} = [{\mathbf{u}^{1}}^T, ..., {\mathbf{u}^{n^{T}}}]^T ,~~~ \mathbf{p} = [{\mathbf{p}^{1}}^{T}, ..., {\mathbf{p}^{n^{T}}}]^T.
\end{equation}
where $n$ is the number of the tubes. Letting $\mathbf{K}_{i}^{3\times3}$ and $\hat{\mathbf{u}}_i(s)$ denote the $3 \times 3$ stiffness matrix and 3D precurvature vector of tube $i$~\cite{rucker2010kinematics}, respectively, similar concatenations are possible to define $\mathbf{K}_{i}$, $\mathbf{\hat{u}}^{i}$, and $\mathbf{\hat{u}}$:
\begin{equation}
    \begin{aligned}
    \textbf{K}_{i} &= \text{diag}(\mathbf{K}_{i}^{3\times3},\ldots,\mathbf{K}_{i}^{3 \times 3}) \in {R^{3N_{i}\times3N_{i}}}, \\
        \mathbf{\hat{u}}^i &= [\mathbf{\hat{u}}_i(s_1)^T, ..., \mathbf{\hat{u}}_i(s_{N_i})^T]^T , ~~ \mathbf{\hat{u}} = [{\mathbf{\hat{u}}^{{1}^T}}, ..., {\mathbf{\hat{u}}^{n^{T}}}]^T.
    \end{aligned}
\end{equation}

% \yc{in equation 4, is it necessary to provide $\mathbf{\hat{u}}$ since it is not used below}
Subsequently, the elastic potential energy of tube $i$ is expressed as
% \jun{Is $\textbf{K}^i$ a typo below?}
\begin{equation} \label{eq:objective_fun}\begin{aligned}
   E_p(\mathbf{u}^i, \mathbf{\hat{u}}^i, \textbf{K}_i) &= \frac{1}{2} (\mathbf{u}^i-\mathbf{\hat{u}}^i)^{T}\textbf{K}_i(\mathbf{u}^i-\mathbf{\hat{u}}^i).
    \end{aligned}
\end{equation}

In the absence of external forces and friction, 
% \yc{do we need to highlight the friction is not considered here, not sure on this}
the shapes of tubes are dominated by the stable equilibrium of the elastic potential energy (i.e., a local minimum). Accordingly, we formulate our large-clearance kinematics model as an energy minimization process subject to tube contact constraints: 
 \begin{subequations} \label{eq:optim_prob}
    \begin{align}
        \underset{\mathbf{u}_i}{\min} ~~ & \sum_{i=1}^{n}E_P(\mathbf{u}_i) \\
         \text{s.t.} ~~ & C(\mathbf{p}_{i-1}, \mathbf{p}_i) \leq \mathbf{0} \text{~for~} i=2,\ldots,n, \\
        & \mathbf{p}_i = F(\mathbf{u}_i) \label{p_u_F}
    \end{align}
\end{subequations}
where $C(\mathbf{p}_{i-1}, \mathbf{p}_i)$ is the contact constraints with which the centerlines of any two adjacent tubes must comply, and $F(\cdot)$ integrates (\ref{p_u_relation}) to calculate $\mathbf{p}_i$ using $\mathbf{u}_i$.

Noting that tube $i$ is the outer tube, tube $i-1$ is the adjacent inner tube, the contact constraint can be derived geometrically by considering the cross-section of tube $i$ at $\mathbf{p}_{i}(s)$, labeled as plane $A$ in Fig. \ref{fig:cross-section}-(b). We remark that, in Fig. \ref{fig:cross-section}-(b), the arc-length locations of the cross-section on the inner and outer tubes are different (i.e., $s_m \neq s_k$) due to the relaxed concentricity. For any $s_{m}$ on the inner tube (i.e., tube $i-1$), a mapping $f_{i-1}(s_m)$ is defined to find the corresponding arc-length position on the outer tube, $s_k$:

\begin{equation} \label{define_f}
    \begin{aligned}
        s_k = f_{i-1}(s_m)= \underset{ \{s_k\} _{, ~k=1,...,N_{i}}}{\text{argmin}} \lVert \textbf{p}_{i}(s_{k}) - \textbf{p}_{i-1}(s_m) \rVert_{_2}
    \end{aligned}
\end{equation}

% \yc{i understand what you mean here, but the second = in the equation doesn't make sense for me, left is the art length $s_k$ can be large, but the right is a minimization and can be very small, the unit also doesn't make sense}\zhouyu{The right expression is used for finding the minimizer for distance in all $s_k$, so the unit should be the same}

The mapping $f_{i-1}$ in (\ref{define_f}) is solved through Nearest Neighbor Search that finds the closest point $\textbf{p}_{i}(s_{k})$ for a given $s_m$. Once $s_k$ is found, the constraint is built on the cross-section perpendicular to the outer tube at $s_k$ in terms of $\textbf{p}_{i}(s_{k})$ and $\textbf{p}_{i-1}(s_{m})$. The constraint $C(\mathbf{p}_{i-1}, \mathbf{p}_i)$ is constituted by collecting the constraint over all $s_{m}$ on the inner tube.

The cross-section of tube $i-1$ at $s_{m}$, after projection into the outer tube's corresponding cross-section, should lie fully inside the latter, as shown in Fig. \ref{fig:cross-section}-(c). The centerline offset $\mathbf{p}_i(s_k) - \mathbf{p}_{i-1}(s_m)$ is projected to plane $A$ as:
\begin{equation} \label{first_order_constraint}
    \begin{aligned}
        \textbf{v} = \textbf{P}_i(s_k) (\textbf{p}_{i}(s_k) - \textbf{p}_{i-1}(s_m))
    \end{aligned}
\end{equation}
where $\textbf{v}$ is the projected offset vector, and $\mathbf{P}_i(s_k)$ denotes the projection matrix of the cross-sectional plane, defined as
\begin{equation} \label{projection matrix}
    \begin{aligned}
        \textbf{P}_{i}(s) = \textbf{I} - \mathbf{p}^{\prime}_{i}(s) \mathbf{p}^{\prime}_{i}(s)^{T}.
    \end{aligned}
\end{equation}
% is the projection matrix that maps any vector in the space into the normal surface of tube $i$ center line at segment $s_{i}$.
%

When the two tubes are locally not parallel at the cross-section, there exists a non-zero  angle $\theta$ between the tubes' tangential directions (see Fig.~\ref{fig:cross-section}-(b)). In this case, the cross-sectional shape of the inner tube is an ellipsoid. By sliding the ellipsoid in contact with the outer tube along the circumference, it is inferred that $\mathbf{p}_{i-1}(s_m)$ is constrained inside an ellipsoidal trajectory $S$, as shown in Fig. \ref{fig:cross-section}-(c). The properties of minor and major axes of $S$ are given by
% Considering the orientation difference $\theta$ between tangential directions of the tube pair, the projected cross-section of inner tube $i-1$ should be an ellipsoid. Following the small increment assumption, the shape deformation and rotation of the ellipsoid during the optimization process is ignored. By sliding the projected inner tube cross-section with fixed rotation, the projected center line position of the inner tube $\mathbf{p}_{i-1}(s_m)$ is constrained inside an ellipsoid trajectory, defined as $S: \mathbf{v}^T \mathbf{Q} \mathbf{v} \leq 1$. From the geometric relations in Fig. \ref{fig:cross-section}-(C), the length and direction of the minor/major axis of the ellipsoid are given by:
\begin{equation}
\label{eq:long_short_axes}
\begin{aligned}
    \mathbf{q}_1 & = \left< \mathbf{P}_i(s_k) \mathbf{p}_{i-1}^{\prime}(s_m) \right >_{_2}, ~~ d_1 = r^{\text{in}}_i - \lambda_1 = r^{\text{in}}_i - r^{\text{out}}_{i-1}  \\
    \mathbf{q}_2 & = \mathbf{q}_1 \times \left < {\mathbf{p}_{i}^{\prime}}(s_k) \right >_{_2}, ~~ d_2 = r^{\text{in}}_i - \lambda_2 = r^{\text{in}}_i - r^{\text{out}}_{i-1} \sec \theta
\end{aligned}
\end{equation} 
% \yc{what is the definition of  $ \mathbf{q}_1$ }
respectively, where $\left< \cdot \right >_{_2}$ denotes the vector normalized w.r.t. $L$-2 norm, $\lambda_1$ and $\lambda_2$ are the lengths of the major and minor axes of the cross-sectional ellipsoid,  {$\mathbf{q}_1$ and $\mathbf{q}_2$ are the unit vectors that represents the direction of the major and minor axes} , $d_1$ and $d_2$ are the lengths of the major and minor axes of ellipsoid $S$,  and $r^{\text{in}}_i$ and $r^{\text{out}}_i$ are the inner and outer radii of tube $i$, respectively. The quadratic equation of $S$ is given as $\mathbf{v}^T \mathbf{Q} \mathbf{v} = 1$, where $\mathbf{Q}$ at tube $i$, segment $s_m$ is calculated by
% Following the quadratic expression of ellipsoids, the analytical form of $\mathbf{Q}$ is given by:
\begin{equation} \label{eq:Q_form}
    \mathbf{Q}_{im} = \frac{1}{d_1^2} \mathbf{q}_1 \mathbf{q}_1^T + \frac{1}{d_2^2} \mathbf{q}_2 \mathbf{q}_2^T.
\end{equation}

Therefore, the constraints on the tube pair can be obtained by combining (\ref{first_order_constraint}) and (\ref{eq:Q_form}) as
% \yc{?? here}
\begin{equation} \label{constrain_p}
     (\textbf{p}_{i}(s_k) - \textbf{p}_{i-1}(s_m))^T ~ \textbf{P}_i^T(s_k) \textbf{Q}_{im} \textbf{P}_i(s_k) ~ (\textbf{p}_{i}(s_k) - \textbf{p}_{i-1}(s_m)) \leq 1
\end{equation}

The energy minimization is non-convex due to the non-convex constraints in (\ref{p_u_F}). By gradually increasing the intertube clearance, we can reduce the induced curvature and position deviation of the inner tube centerline, enabling the linearization of (\ref{p_u_F}). Suppose that, for a given inner radius of outer tube $\Tilde{r}^{\text{in}}_i$, a feasible solution of (\ref{eq:optim_prob}) has already been attained as $\mathbf{\Tilde{p}}^{i}$ and $\mathbf{\Tilde{u}}^{i}$. Then the next sub-problem is defined by imposing a small deviation on clearance $r^{\text{in}}_i = \Tilde{r}^{\text{in}}_i + \Delta c_i$. Assuming small $\Delta \textbf{p}^{i}$ and $\Delta \textbf{u}^{i}$ given a small $\Delta c_i$, a Jacobian mapping $\textbf{J}_i$ can be found to linearize (\ref{p_u_F}) in the neighbourhood of $\mathbf{\Tilde{u}}^{i}$ and $\mathbf{\Tilde{p}}^{i}$. The resulting changes in the centerline and curvature are given by
\begin{equation} \label{Delta_p_u}
\begin{aligned}
    \mathbf{p}^{i} = \Delta \textbf{p}^{i} + \mathbf{\Tilde{p}}^{i}&, 
    ~~  \mathbf{u}^{i} = \Delta \textbf{u}^{i} + \mathbf{\Tilde{u}}^{i} \\
    \Delta \textbf{p}^{i} &= \textbf{J}_i \Delta \textbf{u}^{i} 
    \end{aligned}
\end{equation}
where $\mathbf{J}_{i}$ is obtained by taking derivative of (\ref{p_u_relation}) w.r.t. $\mathbf{u}_i$, the $(k,j)$-th block  $\mathbf{J}_{i}(k,j)$ computed by:
\begin{equation}
\label{how_jacobian_is_calculated}  
    \textbf{J}_{i}(k,j) =\left\{ \begin{aligned}&[\mathbf{\Tilde{p}}_{i}(s_{j})-\mathbf{\Tilde{p}}_{i}(s_{k})]_{\times}\mathbf{R}_{i}(s_{j}) \text{~if~} k \geq j \\
    &\textbf{0} \text{~if~} k < j
    \end{aligned}\right.
\end{equation}

For any adjacent tubes $i$ and $i-1$, $\mathbf{Q}$ is updated with the increased $r^{\text{in}}_i$. Subsequently, we define two pairs of concatenated vectors as 
\begin{equation}
\begin{aligned}
        \Delta \mathbf{u} = [\Delta \mathbf{u}^{{1}^{T}},...,\Delta \mathbf{u}^{{n}^{T}}]^{T}&, ~~~
    \Delta \mathbf{p}= [\Delta \mathbf{p}^{{1}^{T}},...,\Delta \mathbf{p}^{{n}^{T}}]^{T} \\
    \mathbf{\Tilde{u}} = [\mathbf{\Tilde{u}}^{{1}^{T}},...,\mathbf{\Tilde{u}}^{{n}^{T}}]^{T}&, ~~~
    \mathbf{\Tilde{p}}= [ \mathbf{\Tilde{p}}^{{1}^{T}},...,\mathbf{\Tilde{p}}^{{n}^{T}}]^{T} 
\end{aligned}
\end{equation} 

By plugging (\ref{Delta_p_u})-(\ref{how_jacobian_is_calculated}) into (\ref{eq:objective_fun}) and $\ref{constrain_p}$ and ignoring the constant terms, the sub-problem for each increment in $r^{\text{in}}_i$ is formulated as a minimization, of the form
% For any tube pair $i$ and $i-1$, update $Q$ with increased $r^{\text{in}} $ through (\ref{eq:long_short_axes}) and (\ref{eq:Q_form}). Define concatenated vector $ \textbf{u} = [\Delta \textbf{u}_{i-1}^{T},\Delta \textbf{u}_{i}^{T}]^{T}$ and $ \textbf{p} = [\Delta \textbf{p}_{i-1}^{T},\Delta \textbf{p}_{i}^{T}]^{T}$. By plugging (\ref{Delta_p_u})-(\ref{how_jacobian_is_calculated}) into 
 % (\ref{eq:objective_fun}) and $\ref{constrain_p}$ and ignoring the constant terms an optimization problem w.r.t. the change of curvature $\mathbf{u}_i$ is formulated as 
% The objective function is to be optimized with respect to cuvature deviation vector $\Delta u_{1}(s_{i})$ and $\Delta u_{1}(s_{i})$, while the constraints are written with center line position offset $\Delta p_{1}(s_{i})$ and $\Delta p_{2}(s_{i})$. 
% Use the mapping to rewrite the original constraint inequality \ref{boundary_definition_2nd_ord}, and objective function \ref{eq:neat_objective} with respect to concatenated curvature vector $\Delta u = [\Delta u_{1}^{T},\Delta u_{2}^{T}]^{T}$, $u^{*} = [u_{1}(s_{1})^{*T},u_{1}(s_{2})^{*T},...,u_{1}(s_{Ni_{1}})^{*T},u_{2}(s_{1})^{*T},u_{2}(s_{2})^{*T},...,u_{2}(s_{Ni_{2}})^{*T}]^{T}$, $\overset{\wedge}{u} = [\overset{\wedge}{u}(s_{1})^{T},\overset{\wedge}{u}_{1}(s_{2})^{T},...,\overset{\wedge}{u}_{1}(s_{Ni_{1}})^{T},\overset{\wedge}{u}_{2}(s_{1})^{T},\overset{\wedge}{u}_{2}(s_{2})^{T},...,\overset{\wedge}{u}_{2}(s_{Ni_{2}})^{T}]^{T}$
\begin{equation}
    \label{primal_problem}
    \begin{aligned} 
    \min_{\Delta \mathbf{u}} \quad &  \left ( \frac{1}{2}\Delta \mathbf{u}^{T}\mathbf{K}\Delta \mathbf{u} + (\mathbf{\Tilde{u}}-\mathbf{\hat{u}})^{T}\mathbf{K}\Delta \mathbf{u} \right )\\
    \textrm{s.t.} \quad &  \forall m = 1,2,\ldots,N_{i},~~ i = 2,3,\ldots,n\\
    \quad & \frac{1}{2}\Delta \mathbf{u}^{T}\mathbf{X}_{im}\Delta \mathbf{u} + \mathbf{Y}_{im}\Delta \mathbf{u} + \mathbf{Z}_{im} \leq 0, \\
        \quad &\mathbf{X}_{im} = \mathbf{J}_{pi}^{T}\mathbf{S}_{im}^{T}\mathbf{P}_{im}^{T}\mathbf{Q}_{im}\mathbf{P}_{im}\mathbf{S}_{im}\mathbf{J}_{pi}, \\
        \quad &\mathbf{Y}_{im} = \mathbf{\Tilde{p}^{T}}\mathbf{S}_{im}^{T}\mathbf{P}_{im}^{T}\mathbf{Q}_{im}\mathbf{P}_{im}\mathbf{S}_{im}\mathbf{J}_{pi}, \\
        \quad &\mathbf{Z}_{im} = \mathbf{\Tilde{p}^{T}}\mathbf{S}_{im}^{T}\mathbf{P}_{im}^{T}\mathbf{Q}_{im}\mathbf{P}_{im}\mathbf{S}_{im}\mathbf{\Tilde{p}} - 1, \\
        \quad & \mathbf{J}_{pi
        } = \text{diag}(\mathbf{0},\cdots,\mathbf{0},\mathbf{J}_{i-1},\mathbf{J}_{i},\mathbf{0},\cdots,\mathbf{0})\\
        \quad & \mathbf{K} = \text{diag}(\mathbf{K}_{1},...,\mathbf{K}_{n}) \\
        \quad & \mathbf{S}_{im}\mathbf{p} = \mathbf{p}_{i-1}(s_m) - \mathbf{p}_{i}(f_{i-1}(s_{m})) \\
        \quad & \mathbf{P}_{im} = \mathbf{I} - \mathbf{p}_{i}({f_{i-1}(s_{m})}) \mathbf{p}^{\prime}_{i}({f_{i-1}(s_{m})})^{T}.
        % \quad & \mathbf{K} = \scalemath{0.75}{\begin{bmatrix}
        %             \begin{bmatrix}
        %                 \mathbf{K}_{1} & &\\
        %                 & \ddots &\\
        %                 & & \mathbf{K}_{1}
        %                 \end{bmatrix}_{3Ni_{1}\times 3Ni_{1}}
        %              & \mathbf{0} \\
        %             \mathbf{0} & \begin{bmatrix}
        %                 \mathbf{K}_{2} & &\\
        %                 & \ddots &\\
        %                 & & \mathbf{K}_{2}
        %                 \end{bmatrix}_{3Ni_{2}\times 3Ni_{2}}
        %             \end{bmatrix}}
    \end{aligned}
\end{equation}

The problem can be simply viewed as a standard quadratically constrained quadratic programming problem (QCQP) over a variable $\Delta \textbf{u} $. In this letter, the dual problem approach elaborated in \cite{boyd2004convex} is used for solving the QCQP.
\begin{equation}
    \label{dual_problem}
    \begin{aligned}
        \max_{\mathbf{\lambda}} \quad & -\frac{1}{2}\mathbf{q}(\mathbf{\lambda})^{T}\mathbf{P}(\mathbf{\lambda})^{-1}\mathbf{q}(\mathbf{\lambda}) + \mathbf{r}(\mathbf{\lambda}) \\
        \textrm{s.t.} \quad & \forall m = 1,2,...,N_{i}, ~i = 2,3,...,n,~\lambda_{im} \geq 0~ \\
        \quad & \mathbf{P}(\mathbf{\lambda})= \mathbf{K} + \sum_{i,m}\lambda_{im}\mathbf{X}_{im} \\
        \quad & \mathbf{q}(\mathbf{\lambda}) = \mathbf{K}^{T}(\mathbf{\Tilde{u}}-\mathbf{\hat{u}}) + \sum_{i,m}\lambda_{im}\mathbf{Y}_{im}^{T} \\
        \quad & \mathbf{r}(\mathbf{\lambda}) = \sum_{i,m}\mathbf{Z}_{im}
    \end{aligned}
\end{equation}

Specifically, at each step of clearance increment, the dual optimization problem (\ref{dual_problem}), generated from primal problem (\ref{primal_problem}) is optimized with respect to the vector $\mathbf{\lambda}$. This problem comes with simpler constraints. After an optimal $\mathbf{\lambda}$ is reached, $\Delta \textbf{u}$ could be efficiently computed by $\Delta \textbf{u} = \textbf{P}(\mathbf{\lambda})^{-1}\textbf{K}(\mathbf{\Tilde{u}}-\mathbf{\hat{u}})$, since $\Delta \textbf{u}$ is must be the minimizer of Lagrangian. The whole large clearance solver could be formulated iteratively as summarized in Algorithm ~\ref{alg:large_curved}, where the initial condition is obtained using the conventional zero-clearance models \cite{Dupont_2010_CTR_model}.

\begin{algorithm}
\caption{Large clearance algorithm}\label{alg:large_curved}
\begin{algorithmic}
\State \textbf{Initialize} $\mathbf{\hat{u}}$ 
\LineComment{Configure tube pre-curvature}
\State \textbf{Initialize} $\mathbf{\Tilde{p}}$, $\mathbf{\Tilde{u}}$
\LineComment{Zero clearance initial guess from classic CTR models}
\State \textbf{Initialize} $r^{\text{in}}$ $N$ $\Delta c$ 
\LineComment{Configure inner radius, number of incremental steps}
\State $step \gets 1$
\While{$step \leq N
$}
\State Update mapping $f$ from $\mathbf{\Tilde{p}}$
\State Compute Jacobian $\textbf{J}$ from $\mathbf{\Tilde{p}}$
\State $r^{\text{in}} \gets r^{\text{in}}+\Delta c$ 
\State Compute $\mathbf{Q}$ from (\ref{eq:long_short_axes})(\ref{eq:Q_form})
\State Solve (\ref{dual_problem}) for optimal $\lambda$
\State $\Delta \mathbf{u} \gets \textbf{P}^{-1}(\lambda)\mathbf{K}(\mathbf{\Tilde{u}}-\mathbf{\hat{u}})$
\State $ \mathbf{\Tilde{u}} \gets \mathbf{\Tilde{u}}+\Delta \mathbf{u}$
\State Integrate $ \mathbf{\Tilde{u}}$ for updated shape $ \mathbf{\Tilde{p}}$ % \jun{These two lines are linear approximations of $p^*_1(s)$ and $p^*_2(s)$. As step increases, they will accumulate errors. Are they truly updated this way? Or are they calculated by integrating (1a) and (1b) using $u^*_1(s)$ and $u^*_2(s)$?}
\State $step \gets step+1$
\EndWhile
\end{algorithmic}
\end{algorithm}

\subsection{Modeling of CTRs with Impulse Curvature}\label{sec_Large_curvature}

\begin{figure}[t]
  \centering
    \includegraphics[width=0.49\textwidth]{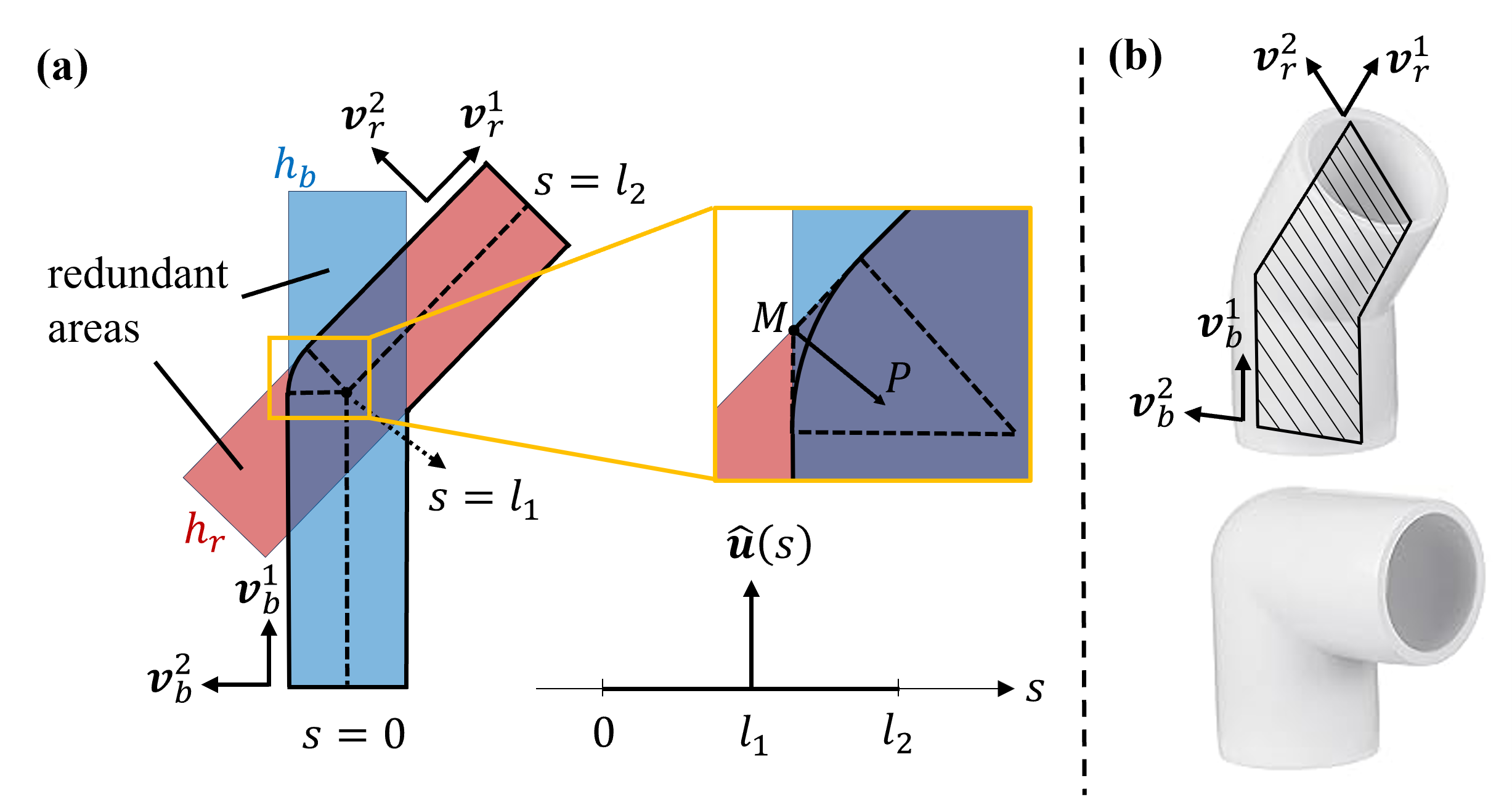}
    \caption{
    Illustration of the tube  with impulse curvature. (a). A schematic diagram that demonstrates the longitudinal section of the elbow joint. The change of the center line curvature with respect to  arc-length along the elbow joint is also displayed. (b). Physical elbow joints used for pipe fittings. The corresponding longitudinal section in (a) is marked out.
    % Elbow joints bending at a acute or obtuse angle, sectioned into 3 parts, each possessing different sets of constraints. Pure red or blue straight section extend along unit directional vector $\textbf{v}_{r}$ and $\textbf{v}_{b}$, while $O$ denotes the center of the overlapping area formed by straight sections. 
    % \yc{remove one sub figure, add a plot showing the curvature vs arc length, which can better describe the impulse curvature assumption we made} 
    }
    \label{fig:elbow_at_different_angles}
\end{figure}

In this section, we will consider a more strict operation scenario when a tube  is manipulated within a channel that has sudden curvature changes at certain places, such as the bifurcation points along the blood vessel or  sharp angles like elbow joints in pipelines. We term this sudden alteration in curvature as ``impulse curvature''. To illustrate, let's consider a complex outer tube that results from the fusion of two straight tubes, painted with red and blue color in Fig. \ref{fig:elbow_at_different_angles}). Curvature remains constant at 0 along red tube, experiences an abrupt transition to infinity at the juncture denoted by $s=l_1$, and then reverts to 0 along blue tube. Analogous to the concept of impulse input in control theory, we categorize this type of tube configuration as having an ``impulse curvature''.
The combined assembly of the inner CTR and outer tube can be approximated as a subset of ``concentric'' tube robots. Notably, the outer tube exhibits significant clearance and possesses impulse curvature.

%These operation channels can be considered as a new class of CTRs, in which the outer tube is sufficiently rigid and has no active DoF. The inner tube, however, is flexible and can be actively translated or rotated within. 
%Let us consider a channel with two straight sections and an elbow joint in between, as illustrated in Fig. \ref{fig:elbow_at_different_angles}, and the outer tube has impulse curvature at $s=l_1$.% Challenges arise when employing our large clearance model in this scenario, as defining the contact constraint at the elbow joint becomes ambiguous.
%Please note that the term ``impulse curvature'' is used in cases where the conventional approach of representing robot shapes with piece-wise constant curvature falls short in providing accurate descriptions. To illustrate, consider the complex outer tube which is formed by joining two straight tubes (labeled as A and B). The curvature along the outer tube is 0 along tube A, becomes infinite at the connection point, and returns to 0 along tube B. Similar to the impulse input in control theory, we classify this kind of tube as having an ``impulse curvature''.

%
% \zhouyu{Let's consider two fundamental instances of such non-smooth channels, as illustrated in Fig. \ref{fig:elbow_at_different_angles}-(a) and \ref{fig:elbow_at_different_angles}-(b). At position $O$, a significant change in channel curvature is observed, transitioning from a finite value to infinity and then returning to a finite value over an infinitesimal arc length. These cases can be regarded as elemental constituents forming more intricate non-smooth channels. Challenges may arise when employing the aforementioned large clearance model in these scenarios, as determining the orientation of the outer tube's cross-section at $O$ becomes intricate, and numerical integration over curvature could lead to compromised performance at points with infinite curvature values.}\yc{zhouyu: from grammar perspective, this is not a sentence. you should avoid this type of description in your paper.}.

To model this type of CTRs, we efficiently view the elbow joint section as being formed by two intersecting straight sections, denoted as the red and blue areas in Fig.~\ref{fig:elbow_at_different_angles}. While straight tubes are considered for presentation simplicity, they can be easily relaxed to non-straight smooth sections that are connected together.
% , with tangential vectors $\mathbf{v}_{r}$ and $\mathbf{v}_{b}$, respectively. We express the projection matrices as follows:
%
% \begin{equation} \label{projection matrix blue red}
%     \begin{aligned}
%         \textbf{P}_{r} = \textbf{I} - \mathbf{v}_{r}{\mathbf{v}_{r}}^{T}, ~~       \mathbf{P}_{b} = \mathbf{I} - \mathbf{v}_{b}{\mathbf{v}_{b}}^{T}
%     \end{aligned}
% \end{equation}
% Finally, the inner tube segment locations are partitioned into three distinct regions: the blue area, the red area, and the overlapping area. We designate three mappings, denoted as $\textbf{R}$,$\textbf{B}$, and $\textbf{T}$, to select from $\textbf{p}$ the blue, red, and overlapping segments.
The inner tube segments are partitioned into three distinct regions: the blue area, the red area, and the overlapping area. We define three selection matrices, $\textbf{B}, \textbf{R}$, and $\textbf{T}$, that leave only the elements of $\textbf{p}$ included in the corresponding regions, respectively, and set the others to zero:

\begin{equation}
    \begin{aligned}
    \label{mapping_written_as_matrices}
        \textbf{B}\textbf{p} & = [\textbf{0},\textbf{0},...,\mathbf{p}(s_{t_1})^T,\textbf{0},...,\mathbf{p}(s_{t_2})^T,...,\textbf{0}]^T_{3N_{i}\times 1} \\
        \textbf{R}\textbf{p} & = [\textbf{0},\textbf{0},...,\mathbf{p}(s_{j_1})^T,\textbf{0},...,\mathbf{p}(s_{j_2})^T,...,,\textbf{0}]^T_{3N_{i}\times 1} \\
        \textbf{T}\textbf{p} & = [\textbf{0},\textbf{0},...,\mathbf{p}(s_{l_1})^T,\textbf{0},...,\mathbf{p}(s_{l_2})^T,...,,\textbf{0}]^T_{3N_{i}\times 1}.
    \end{aligned}
\end{equation}

Let $t,j$ and $l$ represent the indices of inner tube segments situated within the blue, red, and overlapping regions, respectively. The constraints for the red and blue regions can be formulated analogously to (\ref{constrain_p}) as
%
% Let $i$ represents the indices of all inner tube segments situated within the blue only region and their mapped counterparts on outer pipe, $j$ represents those lying within the red only region, and $l$ denotes those situated within the intersection region.
% When considering those portions of the inner tube placed within the regions colored in blue and red, the constraints can be formulated analogously to (\ref{constrain_p}) since it is simply a degenerated case where outer tube curvature is a constant of 0. More precisely, segment $m$ on inner elastic tube is bounded by the contact constraints expressed in (\ref{blue_red_constraints}). Here we follow through the use of concatenated expression of $\textbf{u}$ and $\textbf{p}$:
\begin{equation} \label{blue_red_constraints}
    \begin{aligned}
     h_{b}(\textbf{p}_{1}(s_{t})) & = \textbf{p}^{T}\textbf{S}_{t}^{T}\textbf{P}_{b}^{T}\textbf{Q}\textbf{P}_{b}\textbf{S}_{t}\textbf{p} -1\leq 0, \\
     h_{r}(\textbf{p}_{1}(s_{j})) & = \textbf{p}^{T}\textbf{S}_{j}^{T}\textbf{P}_{r}^{T}\textbf{Q}\textbf{P}_{r}\textbf{S}_{j}\textbf{p} -1\leq 0, \\
      \mathbf{S}_{(\cdot)}\mathbf{p} &= \mathbf{p}_{1}(s_{(\cdot)})-\mathbf{p}_{2}(f_{1}(s_{(\cdot)})),
    \end{aligned}
\end{equation}
where the projection matrices to the cross sections of the blue and red sections are calculated using their tangential vectors $\mathbf{v}_{b}^1$ and $\mathbf{v}_{r}^1$ (see Fig.~\ref{fig:elbow_at_different_angles}-(a)) as
\begin{equation} \label{projection matrix blue red}
    \begin{aligned}
        \mathbf{P}_{b} = \mathbf{I} - \mathbf{v}_{b}^1{\mathbf{v}_{b}^1}^{T}, ~~       \textbf{P}_{r} = \textbf{I} - \mathbf{v}_{r}^1 {\mathbf{v}_{r}^1}^{T}.
    \end{aligned}
\end{equation}

% \zhouyu{In the overlapping area, a potential concern arises when both constraints are simultaneously applied. As the tube shape is calculated iteratively, some segments that were initially within the overlapping area may shift across the borders of sections during the iterations, no longer lying within the corner. However, these segments still satisfy at least one constraint between the two. To address this, we employ the following disjunction constraint in the overlapping area:
% \begin{equation}
%     \begin{aligned}
%     h_{r}(\textbf{p}_{1}(s_{m})) \leq 0 \text{ or } h_{b}(\textbf{p}_{1}(s_{m}))) \leq 0,
%     \end{aligned}
% \end{equation}
% or equivalently,
% \begin{equation} \label{intersection_constraints}
%     \begin{aligned}
%     \min(h_{r}(\textbf{p}_{1}(s_{m})),h_{b}(\textbf{p}_{1}(s_{m}))) \leq 0
%     \end{aligned}
% \end{equation}
% }

In the overlapping area, applying both constraints at the same time can restrict the segments from moving out of the overlapping area during the iterative optimization. To avoid this situation, we consider the following disjunction constraint:
% In the overlapping area, we employ the following disjunction constraint:
\begin{equation}
    \begin{aligned}
    h_{r}(\textbf{p}_{1}(s_{l})) \leq 0 \text{ or } h_{b}(\textbf{p}_{1}(s_{l}))) \leq 0,
    \end{aligned}
\end{equation}
or equivalently,
\begin{equation} \label{intersection_constraints}
    \begin{aligned}
    \min(h_{r}(\textbf{p}_{1}(s_{l})),h_{b}(\textbf{p}_{1}(s_{l}))) \leq 0.
    \end{aligned}
\end{equation}
The feasible area of the above constraint covers all the colored regions in Fig.~\ref{fig:elbow_at_different_angles}-(a), which unintentionally includes unnecessary redundant areas as indicated in the figure. We can cut them out using two planes (with normal vector $\mathbf{v}_b^2$ and $\mathbf{v}_r^2$) tangent to the cylinder of the blue and red section and perpendicular to the longitudinal plane, which gives the following directional constraint:
\begin{equation} \label{eq:cut_redundant}
    \overrightarrow{MP} \cdot \mathbf{v}_b^2 \leq 0, ~~
    \overrightarrow{MP} \cdot \mathbf{v}_r^2 \leq 0,
\end{equation}
where $M$ is a point on the intersection line of the segment planes, $P$ is the point of the center line position $\mathbf{p}_1(s_l)$, and $\mathbf{v}_b^2$ and $\mathbf{v}_r^2$ are the unit normal vectors that represent the segment plane of blue and red section respectively.
% \jun{How are $M$, $\mathbf{v}_b^2$, and $\mathbf{v}_r^2$ defined in 3D case?}

% By applying this constraint to the segments in the overlapping region, we keep the segments from penetrating out of the channel, while letting them freely move across the borders between the regions. \jia{The disjunction constraints (\ref{intersection_constraints}) introduces two redundant areas as shown in Fig. \ref{fig:elbow_at_different_angles}, which can be cut out by applying another directional constraint:
% \begin{equation} \label{eq:cut_redundant}
%     \overrightarrow{MP} \cdot \mathbf{v}_b^2 \leq 0, ~~
%     \overrightarrow{MP} \cdot \mathbf{v}_r^2 \leq 0
% \end{equation}
% where $M$ denotes the intersection point of the segment lines, $P$ is the point of the center line position $\mathbf{p}_1(s_l)$, and $\mathbf{v}_b^2$ and $\mathbf{v}_r^2$ are the unit normal vectors that represent the radial direction of the blue and red section respectively.}

We remark that the outer corner of the elbow joints is usually designed to be rounded in reality (Fig. \ref{fig:elbow_at_different_angles}-(b)) to enhance its durability and prevent the users from safety hazards. This rounded corner is not explicitly modeled since the directional constraint (\ref{eq:cut_redundant}) already makes an approximation of the elbow joint in relatively high precision (Fig. \ref{fig:elbow_at_different_angles}-(a)). Simple modifications could be made to directional constraints for accommodation, should the rounded smooth outer corner be considered. Also, due to the inherent tendency of the inner tube to evolve its shape towards energy minimum, the inner tube normally does not make contact with the rounded outer corner.

Finally, we define $h_c(\mathbf{p}_1(s_l))$ as a combination of the constraints at the corner (\ref{intersection_constraints})-(\ref{eq:cut_redundant}), and an energy minimization in the presence of the elbow joint is formulated as follows:
\begin{equation}\label{elbow_non_linear}
\begin{aligned}
\min_{u} \quad & \frac{1}{2}(\mathbf{u}-\mathbf{\hat{u}})^{T}\textbf{K}(\mathbf{u}-\mathbf{\hat{u}})\\
\textrm{s.t.} % \quad & \forall t \in \{t_1,t_2,...\},j \in \{j_1, j_2,...\},l \in \{l_1,l_2,...\} \\
            \quad & h_b(\textbf{p}_{1}(s_{t})) = h_b(\textbf{B}F(\textbf{u})_{t}) \leq 0,~~ \forall t \in \{t_1,t_2,...\}\\
            \quad & h_r(\textbf{p}_{1}(s_{j})) = h_r(\textbf{R}F(\textbf{u})_{j}) \leq 0, ~~\forall j \in \{j_1, j_2,...\}\\
            % \quad & \min(h_b(\textbf{T}F(\textbf{u})_{l}),h_r(\textbf{T}F(\textbf{u}))_{l}) \leq 0 \\
            \quad & h_c(\mathbf{p}_1(s_l)) = h_c(\mathbf{T} F(\mathbf{u})_l) \leq 0, ~~~ \forall l \in \{l_1,l_2,...\}
\end{aligned}
\end{equation}
%
% This problem is nonlinear because $F(\cdot)$ is a nonlinear function.
% Upon initial inspection, the optimization problem appears intricate, particularly due to the high nonlinearity of the constraints. The computation of $F$ involves a series of calculations for skew-symmetric matrices and matrix exponentials. 

In contrast to the previous case, this problem is not easily reduced to a QCQP due to the non-smooth constraint $h_c(\mathbf{p}_1(s_l))$. Alternatively, we adapt the Sequential Quadratic Programming (SQP), which sequentially approximates the problem as standard QPs and attains incremental solution updates. Each approximation undergoes differentiating constraints, which can be efficiently performed by using the Jacobian mapping in (\ref{how_jacobian_is_calculated}).
% Simplifying the optimization process into more canonical sub-forms, such as QCQP, would prove highly advantageous. Sequential Quadratic Programming (SQP) emerges as a favorable solver in this context. It adopts an iterative approach to optimize variables, focusing primarily on the incremental offset of variables, denoted as $\Delta \textbf{u}$. This permits the use of the Jacobian to map the deviation of $\Delta \textbf{u}$ to the deviation of the constraints. 
With the introduction of Lagrangian multiplier $\lambda$, the corresponding Lagrangain function for this problem could be defined as:
% \begin{equation}
%     \begin{aligned}
%          L(\textbf{u},\lambda) = &E_{P}(\textbf{u}) + \sum_{t} \lambda_{t} (h_{b}((\textbf{B}\textbf{p})_{t})) + \sum_{j} \lambda_{j} (h_{r}((\textbf{R}\textbf{p})_{j})) \\
%                          & + \sum_{l} \lambda_{l} \min(h_{b}((\textbf{B}\textbf{p})_{l}),h_{r}((\textbf{R}\textbf{p})_{l}))
%     \end{aligned}
% \end{equation}
\begin{equation}
    \begin{aligned}
         L(\textbf{u},\lambda) = & ~ E_{P}(\textbf{u}) + \sum_{t} \lambda_{t} (h_{b}((\textbf{B}\textbf{p})_{t})) \\ 
         & + \sum_{j} \lambda_{j} (h_{r}((\textbf{R}\textbf{p})_{j})) 
                          + \sum_{l} \lambda_{l} (h_{c}((\textbf{T}\textbf{p})_{l}))
    \end{aligned}
\end{equation}

For potential energy minimum to be reached, the necessary Karush–Kuhn–Tucker (KKT) conditions must be fulfilled, leading to the solution of equation $\nabla L(\mathbf{u},\lambda) = \mathbf{0}$ through Newton's method. Specifically, at iteration $k$, given a pair of feasible solutions $(\textbf{u}^{k},\lambda^{k})$ and the corresponding $\textbf{p}^{k} ( =F(\textbf{u}^{k}) )$, the SQP subproblem can be formulated as follows: % \jun{I don't understand why $\lambda$ appears here. Please provide me with a reference for the SQP theories.}
\begin{equation}
    \begin{aligned}
    \label{sqp_submin}
        \min_{\textbf{d}} \quad  & E_{P}(\textbf{u}^{k})+\nabla_{\textbf{u}} E_{P}(\textbf{u}^{k})^{T}\textbf{d}+\frac{1}{2}\textbf{d}^{T}\nabla_{\textbf{u}\textbf{u} }^{2}L(\textbf{u}^{k})\textbf{d} \\
        \textrm{s.t.} \quad  & h_b((\textbf{B}\textbf{p}^{k})_{t}) + \nabla_{\textbf{u}}h_b((\textbf{B}\textbf{p}^{k})_{t})^{T}\textbf{d} \leq 0 \\
                        \quad  & h_r((\textbf{R}\textbf{p}^{k})_{j}) + \nabla_{\textbf{u}}h_r((\textbf{R}\textbf{p}^{k})_{j})^{T}\textbf{d} \leq 0 \\
                         & h_c((\textbf{T}\textbf{p}^{k})_{l}) + \nabla_{\textbf{u}}h_c((\textbf{T}\textbf{p}^{k})_{l})^{T}\textbf{d} \leq 0 \\
                        \quad & \textbf{d} = \textbf{u} - \textbf{u}^{k} \\
                        %\quad & L(\textbf{u}^{k}) = E_{P}(\textbf{u}^{k}) + \sum_{t} \lambda^{k}_{t} (h_{b}((\textbf{B}\textbf{p}^{k})_{t})) \\
                        %\quad & \quad \quad  + \sum_{j} \lambda^{k}_{j} (h_{r}((\textbf{R}\textbf{p}^{k})_{j})) +\sum_{l} \lambda^{k}_{l} (h_{c}((\textbf{T}\textbf{p}^{k})_{l}))
    \end{aligned}
\end{equation}
where
\begin{equation}
    \begin{aligned}
    \label{sqp_subproblem}
        \nabla_{\mathbf{u}}h_b(\mathbf{p}^{k}_{m})^{T} &=2{ \mathbf{p}^{k}}^{T}\mathbf{B}^{T}\mathbf{S}_{m}^{T}\mathbf{P}_{b}^{T}\mathbf{Q}\mathbf{P}_{b}\mathbf{S}_{m}\mathbf{B}\mathbf{J}\\
        \nabla_{ \mathbf{u} }h_r(\mathbf{p}^{k}_{m})^{T} &=2{ \mathbf{p}^{k}}^{T}\mathbf{R}^{T}\mathbf{S}_{m}^{T}\mathbf{P}_{r}^{T}\mathbf{Q}\mathbf{P}_{r}\mathbf{S}_{m}\mathbf{R}\mathbf{J}\\
        \nabla_{ \mathbf{u} }h_c(\mathbf{p}^{k}_{l})^{T} &=\mathbf{g}_l + \mathbf{T} \mathbf{J} {\mathbf{V}_{b,l}^2}^T + \mathbf{T} \mathbf{J} {\mathbf{V}_{r,l}^2}^T\\
        \nabla_{ \mathbf{u}\mathbf{u} }^{2}L(\mathbf{u}^{k}) &= \mathbf{K} + \sum_{t}2\lambda_{t}^{k}\mathbf{J}^{T}\mathbf{B}^{T}\mathbf{S}_{t}^{T}\mathbf{P}_{b}^{T}\mathbf{Q}\mathbf{P}_{b}\mathbf{S}_{t}\mathbf{B}\mathbf{J}\\
        &+\sum_{j}2\lambda_{j}^{k}\mathbf{J}^{T}\mathbf{R}^{T}\mathbf{S}_{j}^{T}\mathbf{P}_{r}^{T}\mathbf{Q}\mathbf{P}_{r}\mathbf{S}_{j}\mathbf{R}\mathbf{J}+ \sum_{l}2\lambda_{l}^{k}\mathbf{H}_{l}\\
        \mathbf{J} &= \begin{bmatrix}
            \mathbf{J}_{1} & \mathbf{0} \\
            \mathbf{0} & \mathbf{J}_{2}
        \end{bmatrix}\\
\mathbf{V}^{2}_{b,l} &= [\mathbf{0}_{1\times 3(l-1)},~ {\mathbf{v}_{b}^{2}}^{T},~\mathbf{0}_{1\times 3(N_{i}-l)}] \\
\mathbf{V}^{2}_{r,l} &= [\mathbf{0}_{1\times 3(l-1)},~{\mathbf{v}_{r}^{2}}^{T},~\mathbf{0}_{1\times 3(N_{i}-l) }]
        % &\mathbf{V}^{2}_{b,l} = 
        %     \begin{blockarray}{ccccc}
        %      &  &  & l_{th} &  \\
        %     \begin{block}{[ccccc]}
        %       \mathbf{0}_{1\times3} & \mathbf{0}_{1\times3} & \ldots & {\mathbf{v}_{b}^{2}}^{T} & \mathbf{0}_{1\times3}\\
        %     \end{block}
        %     \end{blockarray}\\
        % &\mathbf{V}^{2}_{r,l} = 
        %     \begin{blockarray}{ccccc} 
        %     &  &  & l_{th} &  \\
        %     \begin{block}{[ccccc]}
        %       \mathbf{0}_{1\times3} & \mathbf{0}_{1\times3} & \ldots & {\mathbf{v}_{r}^{2}}^{T} & \mathbf{0}_{1\times3}\\
        %     \end{block}
        %     \end{blockarray}
    \end{aligned}
\end{equation}

% \yc{there is a $l_{th}$ in the last equation, just want to check this is what you want}\zhouyu{We want to illustrate that the non-zero block vector is placed at the $l_{th}$ slot in the concatented vector.} 

The gradient $\mathbf{g}_l$ and the hessian $\mathbf{H}_l$ of the disjunction constraint (\ref{intersection_constraints}) in above equation are given by:
\begin{equation*}
    \begin{aligned}
        \mathbf{g}_l &=\left\{ 
        \begin{array}{cc}    \nabla_{\mathbf{u}}h_b(\mathbf{p}^{k}_{l})^{T}, &h_b((\textbf{T}\textbf{p}^{k})_{l}) \leq h_r((\textbf{T}\textbf{p}^{k})_{l}) 
        \\ \nabla_{\mathbf{u}}h_r(\mathbf{p}^{k}_{l})^{T},&h_r((\textbf{T}\textbf{p}^{k})_{l}) < h_b((\textbf{T}\textbf{p}^{k})_{l})
        \end{array}\right. \\ 
        \mathbf{H}_{l} &= \left\{ 
        \begin{array}{cc}
    \mathbf{J}^{T}\mathbf{B}^{T}\mathbf{S}_{l}^{T}\mathbf{P}_{b}^{T}\mathbf{Q}\mathbf{P}_{b}\mathbf{S}_{l}\mathbf{B}\mathbf{J},&h_b((\textbf{T}\textbf{p}^{k})_{l}) \leq h_r((\textbf{T}\textbf{p}^{k})_{l})\\
    \mathbf{J}^{T}\mathbf{R}^{T}\mathbf{S}_{l}^{T}\mathbf{P}_{r}^{T}\mathbf{Q}\mathbf{P}_{r}\mathbf{S}_{l}\mathbf{R}\mathbf{J},&h_r((\textbf{T}\textbf{p}^{k})_{l}) < h_b((\textbf{T}\textbf{p}^{k})_{l})
        \end{array}\right.\\
    \end{aligned}
\end{equation*}
after optimal $\mathbf{d}$ is reached, $\lambda$ could be updated by solving:
% \begin{equation}
%     \begin{aligned}
%         \nabla_{\textbf{u}\textbf{u} }^{2}L(\textbf{u}^{k}) + \nabla h^{T}\lambda &= -\nabla E_{P}(\mathbf{u}^{k}) \\
%     h = [h_{b}(\mathbf{p}^{k}_{t}),...,h_{r}(\mathbf{p}^{k}_{j}),&...,\min(h_{b}(\mathbf{p}^{k}_{l}),h_{r}(\mathbf{p}^{k}_{l}))]^T
%     \end{aligned}
% \end{equation}
\begin{gather*}
        \nabla_{\textbf{u}\textbf{u} }^{2}L(\textbf{u}^{k})\mathbf{d} + \nabla h^{T}\lambda = -\nabla E_{P}(\mathbf{u}^{k}), \\
    h = [h_{b}(\mathbf{p}^{k}_{t}),...,h_{r}(\mathbf{p}^{k}_{j}),...,h_{c}(\mathbf{p}^{k}_{l})]^T
\end{gather*}
Note that $\textbf{B}$, $\textbf{R}$, $\textbf{T}$, and $\textbf{S}_{(\cdot)}$ are simply constant matrices in this subproblem.

By employing this approach, the initial nonlinear optimization problem is effectively decomposed into a sequence of standard QPs with linear constraints and quadratic objectives defined in terms of $\textbf{d}$. The resulting $\textbf{d}$ is used to update $\textbf{u}$ by $\textbf{u}^{k+1} = \textbf{u}^{k} + \textbf{d}$. For practical implementation, the numerical optimization can be achieved using \textbf{fmincon} in Matlab with SQP as the underlying algorithm.
% By employing this approach, the initial nonlinear optimization problem is effectively decomposed into a sequence of quadratic programming problems with linear constraints and quadratic objectives defined in terms of $\textbf{d}$, while avoiding the position term $\mathbf{p}$ in the initial constraints (\ref{blue_red_constraints}) by computing the constraints' gradients with respect to $\textbf{u}$. The resulting $\textbf{d}$ is used to update $\textbf{u}$ by $\textbf{u}^{k+1} = \textbf{u}^{k} + \textbf{d}$. For practical implementation, the numerical optimization can be achieved using \textbf{fmincon} in Matlab with SQP as the underlying algorithm.

The iterative radius increment methodology, as previously introduced, can be incorporated together to handle the large clearance and impulse curvature simultaneously. This scenario involves two hierarchical iterative loops, the inner and outer loops of which are the SQP and the iterative radius increment, respectively. For each subproblem of the SQP, the mapping $f$, $\textbf{B}$, $\textbf{R}$, and $\textbf{T}$ are updated prior to solving (\ref{sqp_submin}). This adjustment is essential due to the dynamic evolution of the inner tube's centerline during the SQP.

The method proposed in \cite{ha_2017} is used to provide the initial solution for the very first QP. For this purpose, as the method in \cite{ha_2017} only allows finite tube curvatures, the elbow joint of the outer tube was relaxed into a smooth curve with finite curvature.

\section{Results and Discussions}

\begin{figure} 
    \centering
    \includegraphics[width =\linewidth]{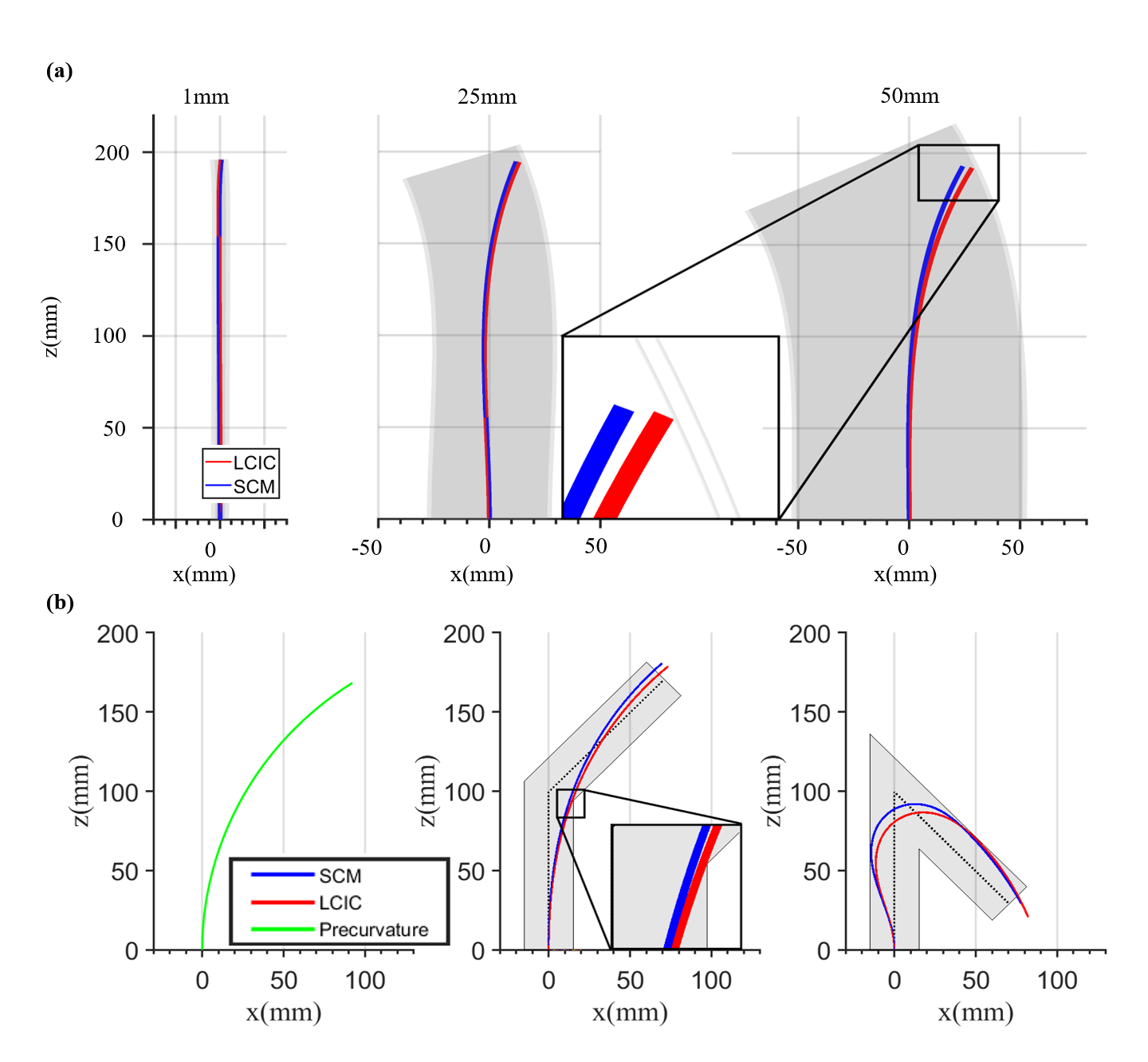}
    \caption{ Simulations of the proposed model at: (a). different outer tube inner radius, (b). different elbow with varying bending angle. In (a) the solution generated by LCIC model makes tip contact with the gray outer tube, while SCM fails to capture such property. In (b), for any chosen bending degree, LCIC model generates more feasible solutions compared to the conventional SCM.}
    \label{fig:simulations}
    \vspace{- 2mm}
\end{figure}

In this section, we will first validate the proposed large clearance and impulse curvature model in numerical simulation. Then we will perform benchtop experiment validations to show the performance of the proposed method and the improvement compared to the state-of-the-art approaches. 

\subsection{Numerical Simulation Studies} \label{numercial_experiments}

Two operation scenarios are considered with different shapes of the outer tube, corresponding to the proposed large curvature model in Sec.\ref{sec_Large_clearance} and the impulse curvature model in Sec.\ref{sec_Large_curvature}:
\begin{enumerate}
\item {\bf Large clearance}. The inner diameter of the outer tube is selected to be $3.32$ mm, $51.32$ mm, and $101.32$ mm in order to represent both small and large clearance scenarios. The inner and outer tubes are subjected to the axial rotation angle of $180^{\degree}$ at the base. 
% \yc{it would be better if we can list the tube properties here as well, i assume you get the tube information from Dr. Ha's paper, then you can cite it to safe some space here} 

\item {\bf Impulse curvature}. Here, we consider the inner tube operating inside a pipe that has an elbow joint of $\{ 45^{\degree}, 135^{\degree} \}$, respectively. 
% the inner tube configuration is taken as the same as the large clearance model. This inner tube is combined with rigid outer pipe elbows, each having a length of $200$ mm and bending at the middle with various angles ($\{ 45^{\degree}, 90^{\degree}, 135^{\degree} \}$). The inner radius of these elbows varied, with options of $10$ mm and $15$ mm. The tube pair is set up with a relative rotation of $0^{\degree}$.
\end{enumerate}
\begin{table}[t!]
\renewcommand{\arraystretch}{1.3}
\caption{Tube Parameters for Numerical Experiments}
\label{table_simu_param}
\centering
\begin{tabularx}{0.5\textwidth}{l |p{0.15\linewidth} l| p{0.15\linewidth} l}
Scenario & \multicolumn{2}{c|}{Large Clearance} &
\multicolumn{2}{c}{Impulse Curvature} \\
\hline % \rule{0pt}{1\normalbaselineskip}
Tube & Outer & Inner & Outer & Inner  \\
\hline % \rule{0pt}{1\normalbaselineskip}
Length [mm]  &  200 & 200 & 200 & 200 \\
Bending Stiffness [Nm$^2$] &  20 & 20 & Inf & 20 \\
Poisson's Ratio  & 0.3 & 0.3 & 0.3 &0.3 \\
Precurvature [m$^{-1}$]  & 0.005 & 0.005 & 0 & 0.005\\
Outer Tube ID [mm] & Var & 0 & 30 & 0 \\
Inner Tube OD [mm] & - & 1.32 & - & 1.32 \\
\hline 
\end{tabularx}
\end{table}
The tube properties used in both scenarios are presented in Table \ref{table_simu_param}. Note that in the impulse curvature scenario, the outer tube is rigid with a shape of elbow joints.

Fig. \ref{fig:simulations}-(a) presents the outcomes of our LCIC model applied to superelastic Nitinol CTRs with a large outer tube, with a comparison to the small clearance model (SCM)  developed in \cite{ha_2017}. With small clearance, both models are able to achieve comparable performance. However, with the increased clearance,  it is clear that our LCIC model is more accurate in predicting the inner tube behavior. Specifically, the distal tip contact is precisely described with the LCIC model, but the SCM fails. In Fig. \ref{fig:simulations}-(b), we showcase the results when a superelastic tube is operating inside a pipe that has an impulse curvature along the body. The proposed model successfully generates feasible configurations subject to various connection angles. Since the impulse curvature scenario in Sec. \ref{sec_Large_curvature} is not considered in \cite{ha_2017}, the SCM simply treats the overlapping segments in (\ref{mapping_written_as_matrices}) as $\mathbf{Tp} = \mathbf{0}_{\{3N_i\times 1\}}$ and removes the corner constraint (\ref{intersection_constraints}) in the optimization problem (\ref{elbow_non_linear}).% \yc{we need to add one sentence after this to show the corresponding consequence of the SCM. the reviewer might ask why the SCM simulation is not shown in Fig. 3}.
% $ $\zhouyu{resolved}
{The resulting inner tube shape generated from SCM in $45^{\degree}$  senario (Fig. \ref{fig:simulations}-(b): middle) is physically unachievable.  It is obvious that the tube is deformed compared to its original shape, which indicates forces/contacts being applied at certain points along the tube, but simulation results indicate no contact between the inner tube and outer channel is observed.

\subsection{Experimental Validation}
\label{hardware_experiments}
In this section, we want to present the experimental validations of our LCIC model with a pre-shaped superelastic Nitinol tube operating at different scenarios. The tube parameters are given in Table \ref{inner_tube_param}.  Both LCIC and SCM models are implemented for comparative study. In all experiment, the accuracy of the models is evaluated from three performance indices: the distal tip error $e_{\text{tip}}$, the average whole-body error $e_{\text{mean}}$, and the maximum whole-body error $e_{\text{max}}$, which are defined as  

%Parameters of the inner tube are given in table \ref{inner_tube_param}. Several experiments are performed to evaluate the accuracy of the CTR model under large clearance. The experiment setups consist of a single tube inserted into different designs of rigid outer pipe. Parameters of the inner tube are given in table \ref{inner_tube_param}.

% We compare the estimated shape of the proposed  with the shape of the SCM. Both models are implemented in Matlab and run on an Intel(R) Core(TM) i7-10875H CPU processor. In all the experiments, the accuracy of the models is evaluated by three factors: the tip error $e_{\text{tip}}$, the average whole-body error $e_{\text{mean}}$, and the maximum whole-body error $e_{\text{max}}$. All errors are calculated from the discrepancies between the model-rendered and the experimental data, defined as
\begin{equation}  \label{error_measure}
\begin{aligned}
    e_{\text{tip}} & = \mathbf{p}_1(s_{N_1}) - \mathbf{p}_{1,\text{ex}}(s_{N_1}) \\
    e_{\text{mean}} & = \frac{1}{N_1} \sum_{i=1}^{N_1} \lVert \mathbf{p}_1(s_i) - \mathbf{p}_{1,\text{ex}}(s_i) \rVert_{_2} \\
    e_{\text{max}} & = \max_{i\in {1,...,N_1}} \lVert \mathbf{p}_1(s_i) - \mathbf{p}_{1,\text{ex}}(s_i) \rVert_{_2}
\end{aligned}
\end{equation}
where $\mathbf{p}_1(s_i)$, $\mathbf{p}_{1,\text{ex}}(s_i)$ denotes the position of the $i$-th discretization point of the inner tube, from the model result and experimental result, respectively. % To handle the fact that the discretization points of models are denser than the manually picked points from the experimental samples, we select a set of points on the model-estimated shape closest to the corresponding points on the experimental shape. 

% Compared with curve-fitting and interpolation of the points on the experimental shape, this approach preserves the features of the original shape. 

\begin{table}[h!]
\renewcommand{\arraystretch}{1.3}
\caption{Parameters of the Inner Tube}
\label{inner_tube_param}
\centering
\begin{tabular}{|c|c|c|c|}
\hline
 ID \textbackslash ~ OD & Total Length & Bending Stiffness  & Poisson's ratio \\
\hline
0.91 \textbackslash ~ 1.32 mm & 213.88 mm & 20.07 Nm$^2$  & 0.3\\
\hline
\end{tabular}
\vspace{- 2mm}
\end{table}

\subsubsection{Planar Scenario}
\label{planar_exp}
We first operate the pre-shaped Nitinol tube inside a 2D channel with a large clearance and a right angle (impulse curvature) (see Fig. \ref{fig:2D_exp_setup}). To simplify the experimental workflow and enable accurate shape measurement of the curved inner tube, the channel is fabricated with 3D printing technique with a thickness equal to the tube diameter, leading to the planar deformation of the inner tube when operating within. 
The channels have three different inner radii $r_2$ of $5$ mm, $10$ mm, and $15$ mm, respectively. For each channel, the inner tube is manually inserted with 3 different insertion depths, measured via a digital calibrator (resolution: 0.1mm). This will result in a total of 9 groups of experimental data. Note that the tube shape in both free- and constrained-condition is captured via camera ($4032 \times 3024$ pixel resolution). To transform the shape data from the camera frame to the global frame, we choose 7 control points with known global positions on the channel (see Fig. \ref{fig:2D_exp_setup}-b)  to perform the coordinate registration (mean registration error: $0.4$ mm).

\begin{figure}[t!]
    \centering
    \includegraphics[width = \linewidth]{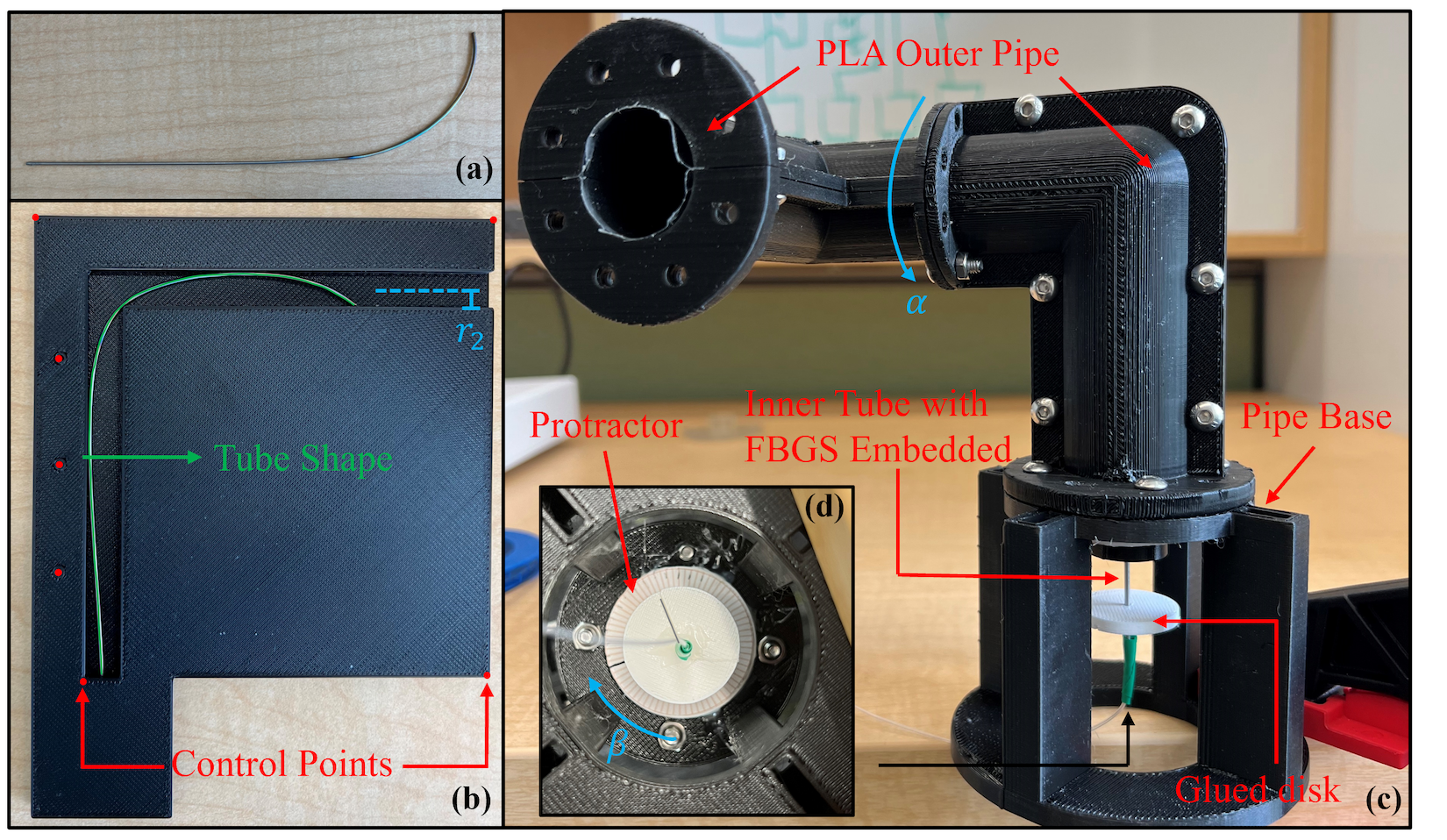}
    \caption{Experiment validations of the CTR clearance model. (a). The original shape of the precurved tube. (b). A planar scenario where a curved tube was inserted into a grooved PLA channel with a inner radi of $r_2$. The control points (red) with known global positions were used for image registration. The shape of the tube inside the outer pipe (green) was extracted by  sampling the points in the image manually. (c). A 3D scenario. The outer channel consisted of two bending segments that could rotate with respect to each other, and a disk was glued to the inner tube to control the orientation. (d). The bottom view of the 3D scenario. The rotation angle of the bending segments and the inne tube is labeled as $\alpha$ and $\beta$ respectively.
    % \yc{the height of this paper can be reduced, also, you can mirror the plots so that it can match Fig 5 below. Lastly, this figure is still too big and takes too much space but the information is minimal}
    }
    \label{fig:2D_exp_setup}
    \vspace{- 2mm}
\end{figure}

% The errors at different operation scenarios for both LCIC and SCM models are shown in Table \ref{table_2D_exp}. It is clear that both models are able to achieve comparable performance when the clearance is small.
% However, the proposed LCIC model is able to reduce the tip error, average body error, and maximum body error when the clearance is large.  Specifically, \yc{please summarize the error value/improvement here. you might need to refer to some papers to see how they write comparative studies. The following description is misleading, you can refer to Bob's publications to get some good samples. this should be done with 2-3 sentences. }

% \textcolor{red}{Jia and Zhouyu: once you address my prior comment, this paragraph can be removed. Errors of tip and whole-body shape predictions for both SCM and  are compared in Table \ref{table_2D_exp}. For $r_2 = 5$ mm that corresponds to $4.34$ mm of clearance, the  produces a tip error of $1.05$ mm, and the average whole-body error of $0.44$ mm, which increases the accuracy of the SCM by $32.2\%$, and $49.4\%$ respectively. Similarly, the  outperforms the SCM by $69.7\%$ of tip error and $72.1\%$ of average whole-body error for $r_2 = 10$ mm, and $80.5\%$ of tip error and $82.8\%$ of average whole-body error for $r_2 = 15$ mm. It can be seen that as the clearance increases, the prediction error of SCM enlarges, while the error of  remains on the same level. The maximum body error of  among all the groups are $2.42$ mm, which corresponds to $1.48\%$ of the total arc-length.}

As shown in Table \ref{table_2D_exp}, the LCIC provides an more accurate estimation of the tube shape than the SCM in all groups of experimental data, especially when the clearance becomes larger. The average tip position error of the LCIC across all groups are $1.53$ mm, which provides an improved accuracy of 71\% compared to the $5.34$ mm average error of the SCM.

The modeled shape and measured shape is detailed in a representative setting that has a large clearance and impulse curvature (see Fig. \ref{fig:PlanarBending_example}). The zoom-in plot in Fig \ref{fig:PlanarBending_example} demonstrates the contact details of the inner tube located at the corner.  Since the centerline of the outer pipe consists of an impulse curvature (90$^\text{o}$ corner), the projection matrix $\mathbf{P}_i(s)$ in (\ref{projection matrix}) changes drastically when the the inner tube enters another section of the elbow joint and therefore the contact constraints (\ref{constrain_p}) shifts. The length of the center line is denoted as $l_m$ for the vertical cylinder. The SCM assumes that the shift of constraints takes place at $s=l_m$, while in reality $s<l_m$ due to the large clearance. This leads to stricter constraints of the SCM at the corner and therefore the failure to model the contact. On the contrary, the disjunction constraint (\ref{intersection_constraints}) of the LCIC fits both the horizontal and vertical section of the elbow joint, and accurately predicts the contact behaviors of the inner tube at the corner.

\begin{figure}[t!]
    \centering
    \includegraphics[width =\linewidth]{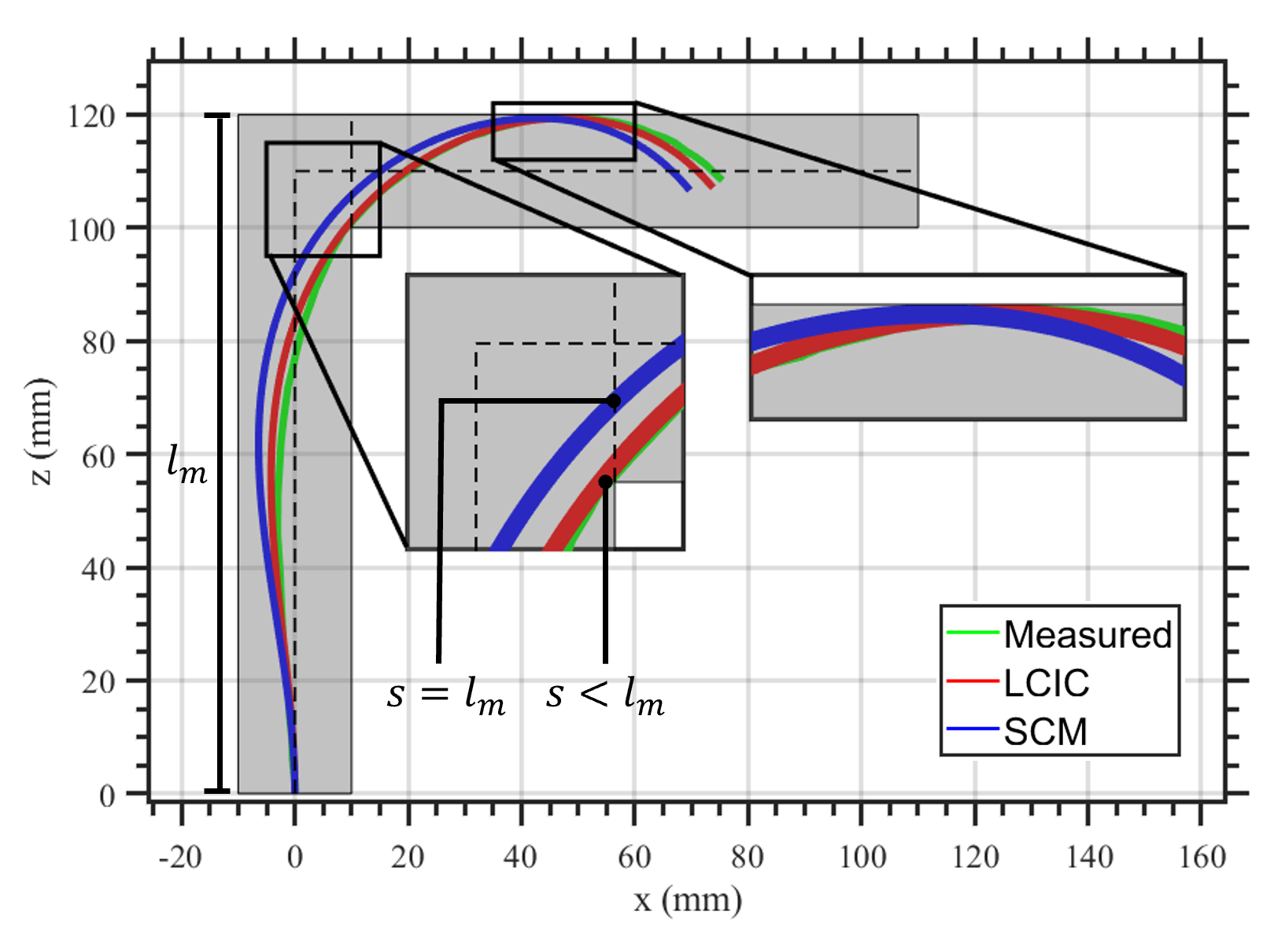}
    \caption{Representative experimental results of a precurved tube in a 90-degree-bent rigid planar pipe (green outline). The result shows a comparison between the shape estimated by SCM, , and the measured shape in experiments. The zoom-in plot near the corner/top region of the outer pipe are also displayed.} 
    % \yc{ from my perspective, Fig. 5 provides more information compared to Fig. 3b, and the figure quality is much better. So do we need to keep Fig. 3b}
    % } 
    \label{fig:PlanarBending_example}
    \vspace{- 4mm}
\end{figure}

% \begin{figure}[t!]
%     \centering
%     \includegraphics[width =\linewidth]{images/plot_rightangle_2D_barplot.png}
%     \caption{Boxplot of error for a precurved tube in a 90-degree-bent rigid planar pipe. The result shows the median error of the shape estimated by the small clearance model (SCM) and the large clearance model (LCM) with respect to the actual shape in the experiment.} 
%     \label{fig:90_deg_exp_barplot}
%     \vspace{- 2mm}
% \end{figure}

% It is important to note that due to potential discrepancies in the discretization of the ground truth shape from measurements and the theoretical shape from simulations, the actual shape error was determined by comparing the position of a segment on the inner tube to the nearest outer pipe segment. To achieve this, a K-nearest neighbors (KNN) search technique was employed, building upon a similar approach presented in \cite{shape_est}.

\begin{table}[t!]
\caption{Error Analysis in 2D Planar Scenario %\yc{the 0.98mm section, why the LCIC is worse than SCM model. I would suggest you double check the data again. Also, we typically put the bad data left column and good data right column. here we need to switch the order}
}
% \label{table_2D_exp}
% \centering
% \begin{tabularx}{0.5\textwidth}{c c|p{0.07\linewidth} l| p{0.07\linewidth} l| p{0.07\linewidth} l}
% \multicolumn{2}{c|}{} & \multicolumn{2}{c|}{$e_{\text{tip}}$ [mm]} &
% \multicolumn{2}{c|}{$e_{\text{mean}}$ [mm]} & \multicolumn{2}{c}{$e_{\text{max}}$ [mm]} \\
% \hline \rule{0pt}{1\normalbaselineskip}
% $r_2$ [mm] & $l_1$ [mm] & LCIC & SCM & LCIC & SCM & LCIC & SCM \\
% \hline \rule{0pt}{1\normalbaselineskip}
%    &  152.76 & 2.17 & 3.56 & 0.66 & 1.09 & 2.17 & 3.56 \\
% 5  &  167.78 & 0.87 & 1.45 & 0.43 & 0.80 & 0.89 & 1.45 \\
%    &  177.42 & 0.75 & 1.81 & 0.34 & 0.78 & 1.04 & 1.62 \\
% \hline \rule{0pt}{1\normalbaselineskip}
%    &  158.38 & 1.54 & 5.38 & 0.61 & 2.62 & 1.54 & 5.38 \\
% 10  &  166.67 & 1.99 & 6.53 & 0.80 & 2.86 & 1.99 & 6.53 \\
%    &  176.93 & 1.86 & 5.95 & 0.85 & 2.61 & 2.13 & 5.95\\
% \hline \rule{0pt}{1\normalbaselineskip}
%    &  157.04 & 0.56 & 6.22 & 0.48 & 3.41 & 0.84 & 6.22 \\
% 15  &  166.76 & 2.42 & 8.67 & 0.78 & 4.14 & 2.42 & 8.67 \\
%    &  176.32 & 1.58 & 8.49 & 0.81 & 4.51 & 1.94 & 8.49 \\
% % \bottomrule
% \hline
% \end{tabularx}
% \vspace{1em}
% % \label{table:2D_exp}
% % \caption{Errors of planar right angle pipe model}
% \vspace{1em}
% \end{table}
\label{table_2D_exp}
\centering
\begin{tabularx}{0.5\textwidth}{c c|p{0.07\linewidth} l| p{0.07\linewidth} l| p{0.07\linewidth} l}
\multicolumn{2}{c|}{} & \multicolumn{2}{c|}{$e_{\text{tip}}$ [mm]} &
\multicolumn{2}{c|}{$e_{\text{mean}}$ [mm]} & \multicolumn{2}{c}{$e_{\text{max}}$ [mm]} \\
\hline \rule{0pt}{1\normalbaselineskip}
$r_2$ [mm] & $l_1$ [mm] & SCM & LCIC & SCM & LCIC & SCM & LCIC \\
\hline \rule{0pt}{1\normalbaselineskip}
   &  152.76 & 3.56 & 2.17 & 1.09 & 0.66 & 3.56 & 2.17 \\
5  &  167.78 & 1.45 & 0.87 & 0.80 & 0.43 & 1.45 & 0.89 \\
   &  177.42 & 1.81 & 0.75 & 0.78 & 0.34 &1.62  & 1.04 \\
\hline \rule{0pt}{1\normalbaselineskip}
   &  158.38 & 5.38 & 1.54 & 2.62 & 0.61 &  5.38 & 1.54\\
10  &  166.67 & 6.53 & 1.99 & 2.86 & 0.80 & 6.53 & 1.99 \\
   &  176.93 & 5.95 & 1.86 & 2.61 & 0.85 & 5.95 & 2.13\\
\hline \rule{0pt}{1\normalbaselineskip}
   &  157.04 & 6.22 & 0.56 & 3.41 & 0.48 & 6.22 & 0.84 \\
15  &  166.76 & 8.67 & 2.42 & 4.14 & 0.78 & 8.67 & 2.42 \\
   &  176.32 & 8.49 & 1.58 & 4.51 & 0.81 & 8.49 & 1.94 \\
% \bottomrule
\hline
\end{tabularx}
\vspace{1em}
% \label{table:2D_exp}
% \caption{Errors of planar right angle pipe model}
\vspace{1em}
\end{table}

\subsubsection{3D  Scenario}
\label{3d_exp}
In this section, we want to evaluate the efficacy of the proposal model in a more complicated operation scenario, where the Nitinol inner tube is manipulated within a 3D channel that has an out-of-plane bending segment (Fig. \ref{fig:2D_exp_setup} shows that an out-of-plane bending segment is perpendicular to the plane that is defined by the other segments). To generate various shapes of the outer channel, the out-of-plane segment is able to rotate relatively with respect to the others and fixed using screws at angles of $\alpha=-45\degree, 0\degree, 45\degree$ (refer to Fig.\ref{fig:90_deg_exp_3D}-(a)-(c). This operation scenario can be found in many engineering applications such as pipe inspection. Plastic tape is applied on the inner wall surface to reduce friction. Due to limited visibility inside the pipe during the experiment, Fiber Bragg Grating (FBG) sensors are chosen instead of imaging systems to measure the shape of the inner tube. Similarly, the inner tube insertion is manually controlled with 4 different lengths, leading to a total of 12 groups of experimental results. A PLA disk is glued to the inner tube, and a protractor fixed at the base is used to measure the tube's rotation angle $\beta$.  
\begin{equation}
    \label{Transform_FBGS_shape}
    [\mathbf{p}_{1,\text{ex}}^T(s), 1]^T = \mathbf{T} ~ [\mathbf{p}_{1, \text{fbg}}^T(s), 1]^T
\end{equation}
The registration is implemented in Matlab using iterative closest point algorithm, and the maximum error is $2.0$ mm. 

The error metrics $e_{\text{tip}}$, $e_{\text{mean}}$, and $e_{\text{max}}$ of both SCM and LCIC models in the 3D experiments are presented in Table \ref{table_3D_exp}. In all groups of experimental results, the prediction of the LCIC shows significant decreased errors compared to that of the SCM. The average tip position error of the LCIC across all groups is $4.36$ mm. Comparing to the $12.64$ mm average error of the SCM, 66\% improvement of the model accuracy is reported. The maximum body error of LCIC among all the groups is $4.90$ mm, which corresponds to $2.73\%$ of the total arc length. Fig. \ref{fig:90_deg_exp_3D} depicts the shapes of inner tube inside the three different 3D pipes corresponding to $\alpha = 45\degree, 0\degree, -45\degree$. In all cases, the shapes predicted by the LCIC model fit well with the experimental data compared to that of the SCM. No contact is observed at the unmodeled outer corners. The LCIC model precisely captures the contact positions at both corners in Fig \ref{fig:90_deg_exp_3D}-(a) because the contact constraints proposed in (\ref{constrain_p}), (\ref{intersection_constraints}), and (\ref{eq:cut_redundant}) are more accurate compared to the existing model.   

% \begin{table}[t!]
% \label{table_3D_exp}
% \caption{Errors of 3D right angle pipe model}
% \centering
% \begin{tabularx}{0.5\textwidth}{c c|p{0.07\linewidth} l| p{0.07\linewidth} l| p{0.07\linewidth} l}
% \multicolumn{2}{c|}{} & \multicolumn{2}{c|}{$e_{\text{tip}}$ [mm]} &
% \multicolumn{2}{c|}{$e_{\text{mean}}$ [mm]} & \multicolumn{2}{c}{$e_{\text{max}}$ [mm]} \\
% \hline \rule{0pt}{1\normalbaselineskip}
% $\alpha$ [\degree] & $l_1$ [mm] &  & SCM &  & SCM &  & SCM \\
% \hline \rule{0pt}{1\normalbaselineskip}
%     &  176.38 & 3.18 & 16.98 & 1.97 & 6.45 & 4.18 & 16.97 \\
% 45  &  179.52 & 3.06 & 19.72 & 2.15 & 6.56 & 4.90 & 19.72\\
%     &  182.89 & 2.85 & 16.63 & 1.47 & 6.38 & 2.92 & 16.63 \\
%     &  184.98 & 3.4 & 14.53 & 1.81 & 5.08 & 4.03 & 14.53 \\
% \hline \rule{0pt}{1\normalbaselineskip}
%      &  176.38 & 4.94 & 10.76 & 2.05 & 4.61 & 4.23 & 9.97\\
% -45  &  179.52 & 4.95 & 6.75 & 1.64 & 3.62 & 4.76 & 6.77\\
%      &  182.89 & 4.59 & 11.65 & 1.22 & 3.26 & 3.79  & 11.63\\
%      &  184.98 & 5.46 & 7.65 & 1.17  & 2.98 & 3.74 & 7.63\\
% \hline \rule{0pt}{1\normalbaselineskip}
%    &  176.38 & 5.35 & 11.71 & 1.98 & 4.08 & 3.97 & 11.13\\
% 0  &  179.52 & 4.16 & 13.89 & 1.73 & 4.34 & 3.56 & 13.51\\
%    &  182.89 & 5.36 & 10.75 & 1.80 & 4.38 & 4.03 & 10.50\\
%    &  184.98 & 4.98 & 10.61 & 1.54 & 3.92 & 4.11 & 10.61\\
% % \bottomrule
% \hline
% \end{tabularx}
% \vspace{1em}
% \end{table}

\begin{figure*}[t!]
    \centering
    \includegraphics[width =\linewidth]{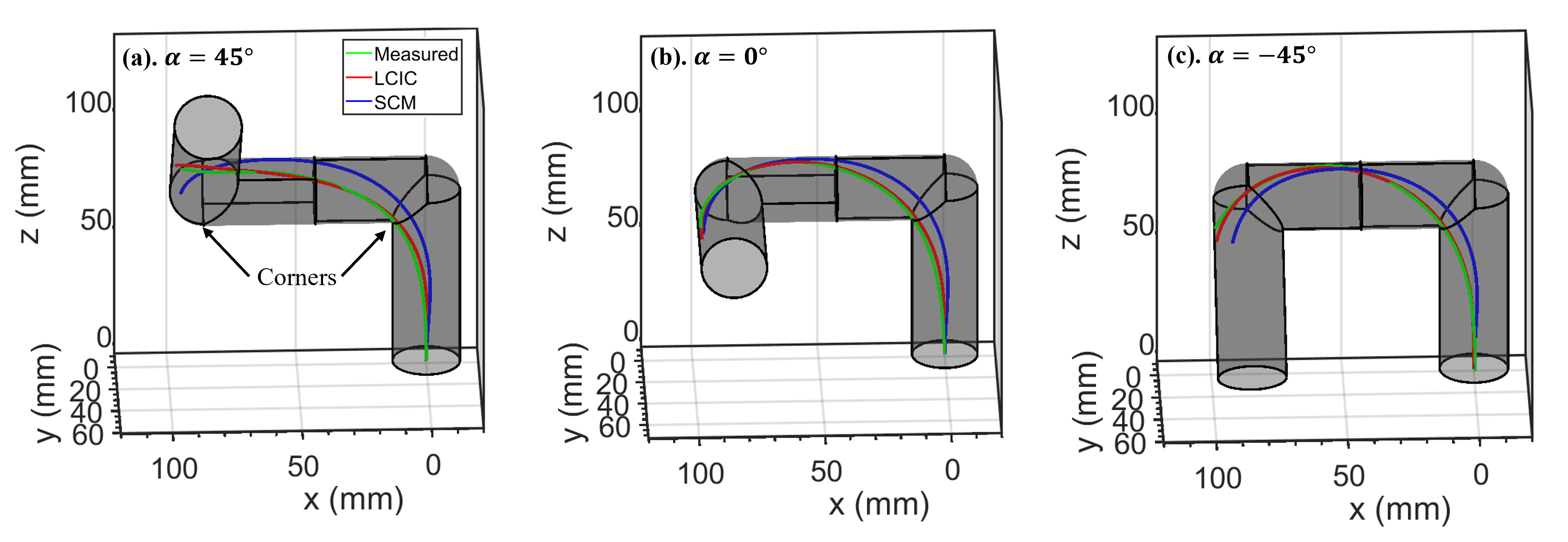}
    \caption{Representative experimental results of a precurved tube in a rigid 3D pipe with 90$^o$ elbow joints. The result shows a comparison between the shape estimated by SCM, LCIC, and the measured shape in experiments.
    } 
    \label{fig:90_deg_exp_3D}
    % \vspace{- 2mm}
\end{figure*}

\begin{table}[t!]
\caption{Errors of 3D right angle pipe model}
\label{table_3D_exp}
\centering
\begin{tabularx}{0.5\textwidth}{r r p{0.08\linewidth} |p{0.055\linewidth} r| p{0.055\linewidth} r| p{0.055\linewidth} r}
\multicolumn{3}{c|}{} & \multicolumn{2}{c|}{$e_{\text{tip}}$ [mm]} &
\multicolumn{2}{c|}{$e_{\text{mean}}$ [mm]} & \multicolumn{2}{c}{$e_{\text{max}}$ [mm]} \\
\hline \rule{0pt}{1\normalbaselineskip} 
$\alpha$~[\degree] & $\beta$~[\degree] & $l_1$~[mm] & SCM & LCIC  & SCM & LCIC & SCM & LCIC \\
\hline \rule{0pt}{1\normalbaselineskip}
45& 150  & 176.38  & 16.98 & 3.18  & 6.45 & 1.97  & 16.97 & 4.18\\
45& 150  & 179.52  & 19.72 & 3.06  & 6.56 & 2.15  & 19.72 & 4.90\\
45& 150  & 182.89  & 16.63 & 2.85  & 6.38 & 1.47  & 16.63 & 2.92\\
45& 150  & 184.98  & 14.53 & 3.40  & 5.08 & 1.81  & 14.53 & 4.03 \\
\hline \rule{0pt}{1\normalbaselineskip}
-45  & 85  & 176.38  & 10.76 & 4.94  & 4.61  & 2.05   & 9.97   & 4.23\\
-45  & 85  & 179.52  & 6.75  & 4.95  & 3.62  & 1.64   & 6.77   & 4.76  \\
-45  & 85  & 182.89  & 11.65 & 4.59  & 3.26  & 1.22   & 11.63  & 3.79 \\
-45  & 85  & 184.98  & 7.65  & 5.46  & 2.98  & 1.17   & 7.63   & 3.74   \\
\hline \rule{0pt}{1\normalbaselineskip}
0  & 115  & 176.38  & 11.71 & 5.35 & 4.08 & 1.98 & 11.13 & 3.97 \\
0  & 115  & 179.52  & 13.89 & 4.16 & 4.34 & 1.73 & 13.51 & 3.56 \\
0  & 115  & 182.89  & 10.75 & 5.36 & 4.38 & 1.80 & 10.50 & 4.03 \\
0  & 115  & 184.98  & 10.61 & 4.98 & 3.92 & 1.54 & 10.61 & 4.11 \\
% \bottomrule
\hline
\end{tabularx}
\vspace{1em}
\end{table}

% error source: friction? sensing error
% The unmodeled friction, though reduced by the inner wall tape, is likely a significant effect since it allows the inner tube to slide inside the pipe near the equilibrium. This phenomena is able to account for the large tip errors ($\sim$5mm) compared to the mean body error ($\sim$2mm), because small deflections at the proximal positions, i.e. corner points, may still lead to large errors at the distal tip. The mean registration error is 2.0mm, which encompasses the measurement error of tube rotation/translation, integration error of FBG sensor, and clearance between the pipe base and inner tube. Although the resolution of the protractor is $5 \degree$, the rotation angles of tube in all groups are designed to be the multiplier of $5$ so that no estimation is needed.  

\section{Conclusion} \label{conclusion}
In this letter, we present the CTR robot model that can accommodate the large clearance and impulse curvature. The proposed method demonstrates increased accuracy compared to the existing small clearance model in both 2D planar and 3D operation scenarios, leading to a tip position error of $1.53$ mm and $4.36$ mm,  outperforming the state-of-the-art  by 71\% and 66\%, respectively. 
This proposed LCIC model makes one step further to the precise modeling of CTRs, with the prospect for applications  inside large channels such as  aorta, lung bronchi, and industrial pipes.   % We also expect this work provides a valuable start point for path planning algorithms enabling CTRs to smartly utilize the mechanics and obstacles of the outer world as supports to reach its target.

\ifCLASSOPTIONcaptionsoff
  \newpage
\fi


% \clearpage
\bibliographystyle{ieeetr}
\bibliography{reference}

\end{document}